\documentclass[10pt,twocolumn,letterpaper]{article}
\usepackage{booktabs}
\usepackage{cvpr}              

\usepackage{graphicx}
\usepackage{amsmath}
\usepackage{amssymb}
\usepackage{booktabs}
\usepackage{multirow}
\usepackage{subcaption}
\usepackage{enumitem}
\usepackage[table,xcdraw]{xcolor}

\usepackage{algorithm}
\usepackage{algpseudocode}

\usepackage[pagebackref,breaklinks,colorlinks]{hyperref}

\usepackage[capitalize]{cleveref}
\crefname{section}{Sec.}{Secs.}
\Crefname{section}{Section}{Sections}
\Crefname{table}{Table}{Tables}
\crefname{table}{Tab.}{Tabs.}

\def\cvprPaperID{6638} 
\def\confName{CVPR}
\def\confYear{2022}

\begin{document}

\title{Attentive Prototypes for Source-free Unsupervised \\Domain Adaptive 3D Object Detection}

\author{Deepti Hegde\\
Johns Hopkins University\\
{\tt\small dhegde1@jhu.edu}
\and
Vishal M. Patel\\
Johns Hopkins University\\
{\tt\small vpatel36@jhu.edu}
}
\maketitle

\begin{abstract}
   3D object detection networks tend to be biased towards the data they are trained on. Evaluation on datasets captured in different locations, conditions or sensors than that of the training (source) data results in a drop in model performance due to the gap in distribution with the test (or target) data. Current methods for domain adaptation either assume access to source data during training, which may not be available due to privacy or memory concerns, or require a sequence of lidar frames as an input. We propose a single-frame approach for source-free, unsupervised domain adaptation of lidar-based 3D object detectors that uses class prototypes to mitigate the effect pseudo-label noise. Addressing the limitations of traditional feature aggregation methods for prototype computation in the presence of noisy labels, we utilize a transformer module to identify outlier ROI's that correspond to incorrect, over-confident annotations, and compute an \textbf{attentive class prototype}. Under an iterative training strategy, the losses associated with noisy pseudo labels are down-weighed and thus refined in the process of self-training.  To validate the effectiveness of our proposed approach, we examine the domain shift associated with networks trained on large, label-rich datasets (such as the Waymo Open Dataset and nuScenes) and evaluate on smaller, label-poor datasets (such as KITTI) and vice-versa. We demonstrate our approach on two recent object detectors and achieve results that out-perform the other domain adaptation works. Code is available at \url{https://github.com/deeptibhegde/AttentivePrototypeSFUDA}
\end{abstract}
\vspace{-1em}

\section{Introduction}
\label{sec:intro}
\begin{figure}
    \centering
    \includegraphics[width=0.8\linewidth]{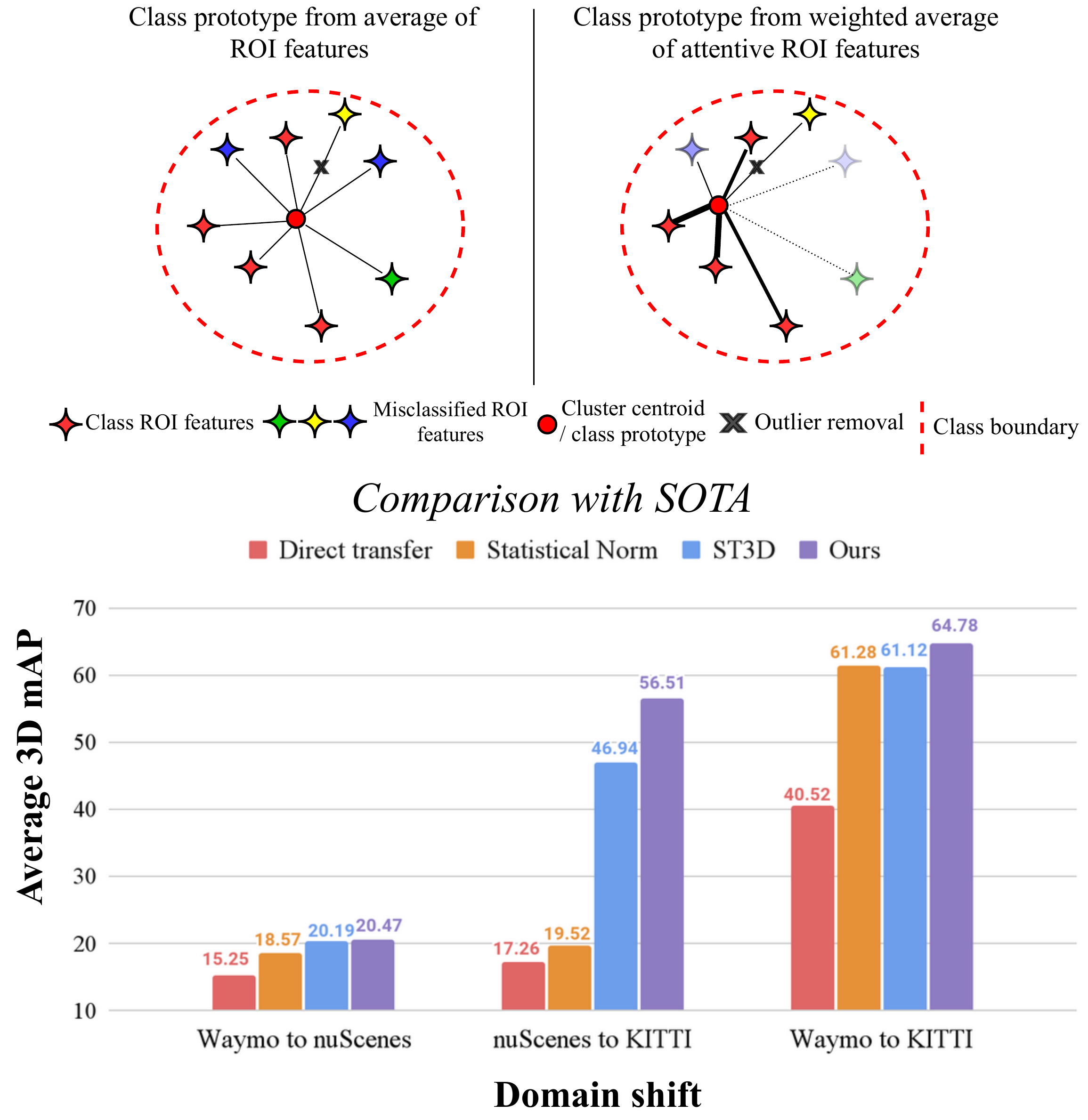}
    \caption{ \textbf{Top row:} Visual representations of prototype computation. On the left hand side is a depiction of a standard feature aggregation approach for prototype computation. In the case of noisy labels, features corresponding to mis-labeled regions that are not discarded by outlier removal contribute to the class prototype. On the right is the proposed method of entropy-weighted average of attentive region features which considers only salient regions for prototype computation. The opacity of the features represents the attention weights, and the width of the connecting lines represents the combination weights for computing the average. \textbf{Bottom row:} Comparison of our results on SECOND-iou against recent state-of-the-art methods for three domain shift scenarios.  }
    \vspace{-1em}
    \label{fig:abs}
\end{figure}
The localization and categorization of objects in a 3D scene is a crucial component of perception systems in fields like robotics and autonomous driving. In recent years, data-driven approaches using deep neural networks have achieved superior performance in various versions of this task \cite{shi2019pointrcnn,second,shi2020pv,Shi2019PartA2N3,REF:qi2017pointnet,ipod,zhou2018voxelnet}, facilitated in part by the release of numerous datasets and benchmarks \cite{waymo,KITTI,nuscenes2019,lyft,chang2019argoverse,cadc}. In practical scenarios, it is important for these object detection frameworks to perform consistently well in different domain scenarios. However, deep neural networks tend to learn not only the valuable features that aid in performing the task at hand, but also the biases present in the data it is trained on. In the case of lidar datasets, the weather conditions and the location of capture lead to biases in the dataset due to the specific dimensions of roads, vehicles, and the driving conventions of the area. Additionally, different lidar sensors possess different rates of return and produce pointclouds with varying densities, leading to another set of inherent biases.  This leads to a distribution gap among various pointcloud datasets. 
\begin{figure}
    \centering
    \includegraphics[width=\linewidth]{ 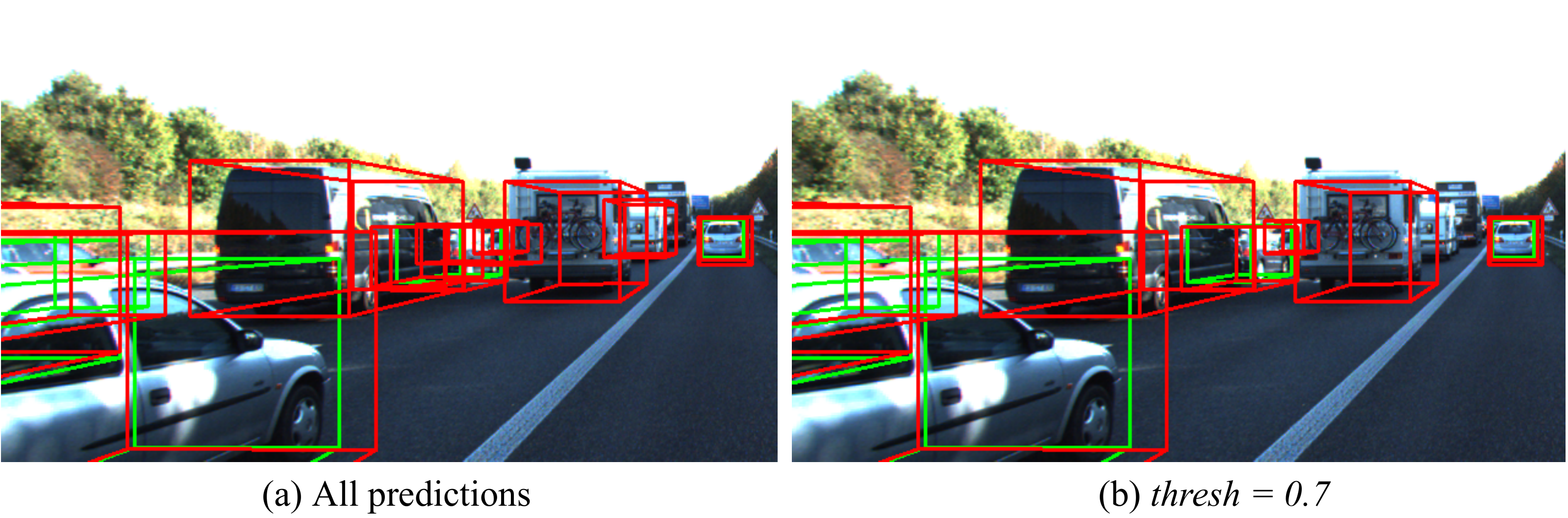}
    \caption{3D bounding box predictions of the object detector  \cite{shi2019pointrcnn} trained on Waymo \cite{waymo} data and tested on KITTI \cite{KITTI}. Ground truth annotations are in green and predictions are in red. Thresholding (b) fails to remove all false positives present in (a). Self-training with these pseudo-labels leads to the enforcement of errors.
    }
    \label{fig:compare_bb}
    \vspace{-1em}
\end{figure}
Thus, an object detector trained on a particular dataset will drop in performance when evaluated on samples from a dataset with a different distribution. We call the training and test datasets in this scenario as the source and target domain datasets, respectively. One may argue that making use of a large, diverse source domain could solve this problem, however there will always be samples from an unseen distribution, and collecting every possible type of lidar scene is impractical at best.  

Unsupervised domain adaptation (UDA) refers to the process of bridging this gap to improve the performance of source-trained networks on unlabelled target samples. There have been several recent works addressing this problem, for both 2D \cite{oza2021unsupervised,RoyChowdhury,Khodabandeh2019ARL} and 3D \cite{xu2021spg,yang2021st3d,Saltori2020SFUDA3DSU,scalablePseudo,Luo2021multiLevel} object detection. However, source samples are often unavailable during training due to limited memory capabilities or privacy reasons. This requires a source-free domain adaptation approach, where only the source-trained model and unlabelled target samples are used for adaptation. There exist several such approaches for various computer vision tasks on images \cite{Li2021AFL,Kundu2020UniversalSD,yang2021generalized}. Only SF-UDA\textsuperscript{3D} \cite{Saltori2020SFUDA3DSU} attempts this setting for 3D object detection, but relies on a sequence of lidar frames as an input to the network.

Self training-based methods have been successful in unsupervised and semi-supervised domain adaptive works \cite{scalablePseudo,xie2020self,zou2019confidence}, but rely on confidence thresholding to filter noisy pseudo-labels. As illustrated in the example from Figure \ref{fig:compare_bb}, the use of high thresholds (as is general practice) results in training the model on easy samples and incorrect labels of high confidence that contribute to the enforcement of errors during adaptation.

   We propose an unsupervised, source-free domain adaptation framework for 3D object detection that addresses the issue of incorrect, over-confident pseudo labels during self-training through the use of class prototypes. In the presence of label noise, standard feature aggregation methods of prototype computation \cite{ZhangProto,lvq,yang2018robust,jiang2018learning} are insufficient, since features corresponding to incorrectly labeled regions could contribute to the final prototype (see Figure \ref{fig:abs}, top row). Inspired by the high representative power of self-attention and recent works that make use of transformers to focus on salient inputs \cite{vit,vaswani2017attention}, we calculate an attentive class prototype by using a transformer to identify salient regions-of-interest and combine their associated feature vectors using prediction entropy weights that represent the uncertainty of the classification branch for each sample.  Class predictions corresponding to incorrect pseudo labels, which are identified by calculating the similarity with the class prototype, are down-weighed to prevent reinforcing errors during self-training. We demonstrate our result on several domain shift scenarios (see Figure \ref{fig:abs}).  Our contributions are as follows  

\begin{itemize}[topsep=0pt,noitemsep,leftmargin=*]
    \item  We propose the \textbf{attentive prototype} for learning representative class features in the presence of label noise by leveraging self-attention through a transformer block and perform source-free unsupervised domain adaptation of 3D object detection networks that mitigates the effect of label noise during self training by filtering incorrect annotations. 
    \item We demonstrate our method on two recent object detectors, SECOND-iou \cite{second}, and PointRCNN \cite{shi2019pointrcnn} for six domain shift scenarios and outperform recent domain adaptation works.
\end{itemize}

\section{Related Works}

\noindent\textbf{3D object detection.} With the increase in availability of multiple large-scale lidar datasets, there have been many recent networks proposed for 3D object detection. Here, we focus on pure lidar-based detectors, although there have been several successful multi-modal approaches \cite{chen2017multi,multivox,ku2018joint}. The seminal works PointNet \cite{REF:qi2017pointnet} and PointNet++ \cite{REF:qi2017pointnetplusplus} for the hierarchical feature extraction of pointclouds have spurred numerous deep neural networks for this task that can be broadly categorised as voxel-based methods \cite{zhou2018voxelnet,lang2019pointpillars,second}, which divide the pointcloud into volumetric grids before performing feature extraction,  and point-based methods \cite{shi2019pointrcnn,yang20203dssd} for 3D object detection, which operate directly each point in the 3D scene. In \cite{second}, Yan \etal propose SECOND, a single stage, voxel-based method that utilizes 3D sparse convolutions and a Region Proposal Network (RPN) head to predict the location and category of objects in a lidar scene. Pointpillars \cite{lang2019pointpillars} by Lang \etal builds on this by changing the shape of the voxels to columns that span the height of the pointcloud. PointRCNN is a two-stage network that generates 3D bounding box proposals followed by a refinement stage similar to \cite{fasterrcnn}. 

\noindent\textbf{Self-training for domain adaptation.}  Pseudo label-based self-training is a popular approach for unsupervised domain adaptation. In \cite{RoyChowdhury}, RoyChowdhury \etal propose a method that trains the object detection network FasterRCNN \cite{fasterrcnn} with a combination of high confidence pseudo labels and labels obtained from a tracker and a knowledge distillation loss to adapt the network for face detection. This approach relies on video data to obtain its tracking results. Khodabandeh \etal \cite{Khodabandeh2019ARL} propose Robust FasterRCNN, a domain adaptive 2D object detector that consists of a source-model training stage, a pseudo label generation stage using a pretrained image classifier for refinement, and a final stage where the detector network is trained on the source labels and the refined target pseudo-labels.
Yang \etal put forth ST3D \cite{yang2021st3d}, a self training approach for 3D domain adaptive object detection where the network is adapted by training with a proposed curriculum data augmentation algorithm using pseudo-labels generated with a quality-aware memory bank. While showing promising results in some cases, in others this method is outperformed by standard pseudo-label based self-training, and depends on boosting with statistical normalization \cite{Wang2020TrainIG}, a weakly supervised method, for its best reported results. Caine \etal \cite{Caine} propose a simple, yet effective domain adaptation method for 3D object detection using the base network of Pointpillars \cite{lang2019pointpillars} that trains a student network with a combination of source labels and target pseudo labels obtained from a teacher network trained on the labeled source data. However, simple thresholding methods for pseudo-label collection from the source model may lead to training the model with incorrect labels that have high confidence and discarding correct labels with confidence that falls below the threshold. 

\noindent\textbf{Source-free domain adaptation.} Privacy and memory limitations during adaptation may prevent the access of source data for training. Source-free approaches only use unlabeled target data network models pre-trained on the source data during adaptation. Kundu \etal \cite{Kundu2020UniversalSD} propose a UDA method which does not require information about the category-level gap between domains, and consists of a procurement stage in which a generative classifier equips the model to reject out-of-distribution samples, and a deployment stage, in which adversarial alignment is performed. Yang \etal \cite{yang2021generalized} focuses on maintaining model performance on the source domain by a two stage method that consists of a local structure clustering stage and a domain attention stage that activates feature channels associated with each domain. Saltori \etal \cite{Saltori2020SFUDA3DSU} propose a source-free UDA method for 3D object detection on PointRCNN \cite{shi2019pointrcnn}, utilizing a tracking-based scoring system to evaluate the quality of pseudo-labels at different scales. As mentioned above, this method depends on the use of multiple frames for a single forward pass through the network.

\noindent\textbf{Prototype learning.} Learning representative features of a class or group of samples has been a well explored problem in pattern recognition. Originally calculated with hand-crafted features\cite{lvq}, recent approaches use convolutional neural networks for feature extraction. This method has seen success in a variety of tasks, including classification \cite{yang2018robust}, zero-shot recognition \cite{jiang2018learning}, and domain adaptive 2D segmentation \cite{ZhangProto}. We argue that when training a network with noisy pseudo-labels, clustering and outlier removal are insufficient in obtaining a true representative class prototype. We thus propose a transformer-based approach to generate attentive prototypes. 


\begin{figure*}
    \centering
    \includegraphics[width=0.8\textwidth]{ 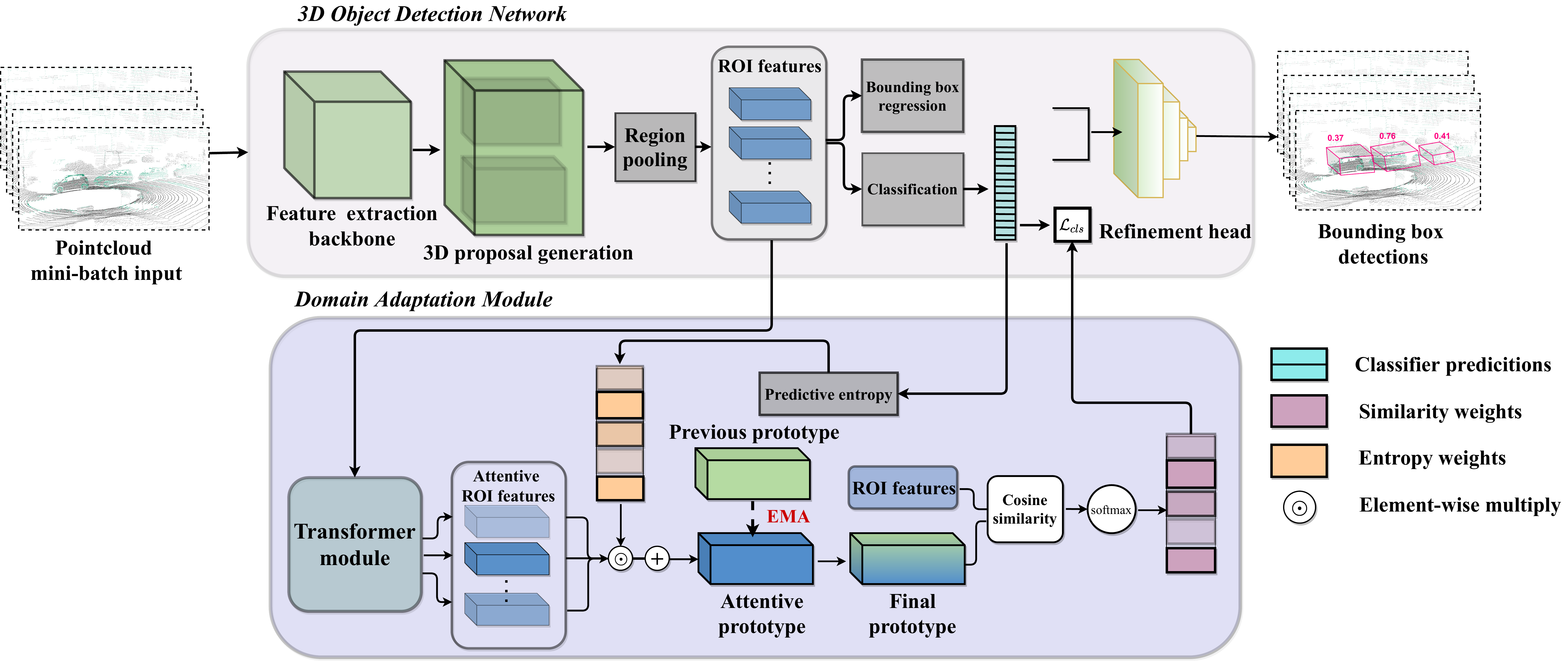}
    \caption{A visual overview of the proposed domain adaptation framework. The object detector is initialised with the source-trained model and used for region-proposal and feature extraction by the domain adaptation block. This consists of a transformer module to generate attentive features (see Figure \ref{fig:trans} for details), prototype computation by combination of features using predictive entropy weights and the exponential moving average (EMA) algorithm, and calculation of the output classification loss weights using cosine similarity. The detection network is trained under the weighted loss.  }
    \label{fig:main}
    \vspace{-1em}
\end{figure*}
\section{Proposed Method}
Consider an object detector network $\phi_s$ trained on a source dataset consisting of $N$ sample-label pairs $\{X^S,Y^S\}=\{x^{S}_i,y^{S}_i\}_{i=1}^N$, where $Y^S$ contains the annotations of objects in a 3D scene, consisting of the dimensions $\{l,w,h\}$, position $\{x,y,z\}$ and category of each bounding box. We aim to adapt this network model in absence of the source data to an unlabeled target dataset $X^T=\{x^T_j\}_{j=1}^M$ of size $M$ and corresponding pseudo-labels $Y^{Ps}=\{y^{Ps}_j\}_{j=1}^M$ generated by $\phi_s$. The proposed domain adaptation method consists of a prototype computation and similarity-based refinement, implemented with an iterative training strategy. A visual representation of this framework is shown in Figure \ref{fig:main}, and a detailed algorithmic description is given in Algorithm \ref{alg:cap}.

\label{sec:trans}
\subsection{Transformer for prototype computation}
\begin{figure}
     \centering
     \begin{subfigure}[b]{0.7\linewidth}

         \includegraphics[width=\linewidth]{ 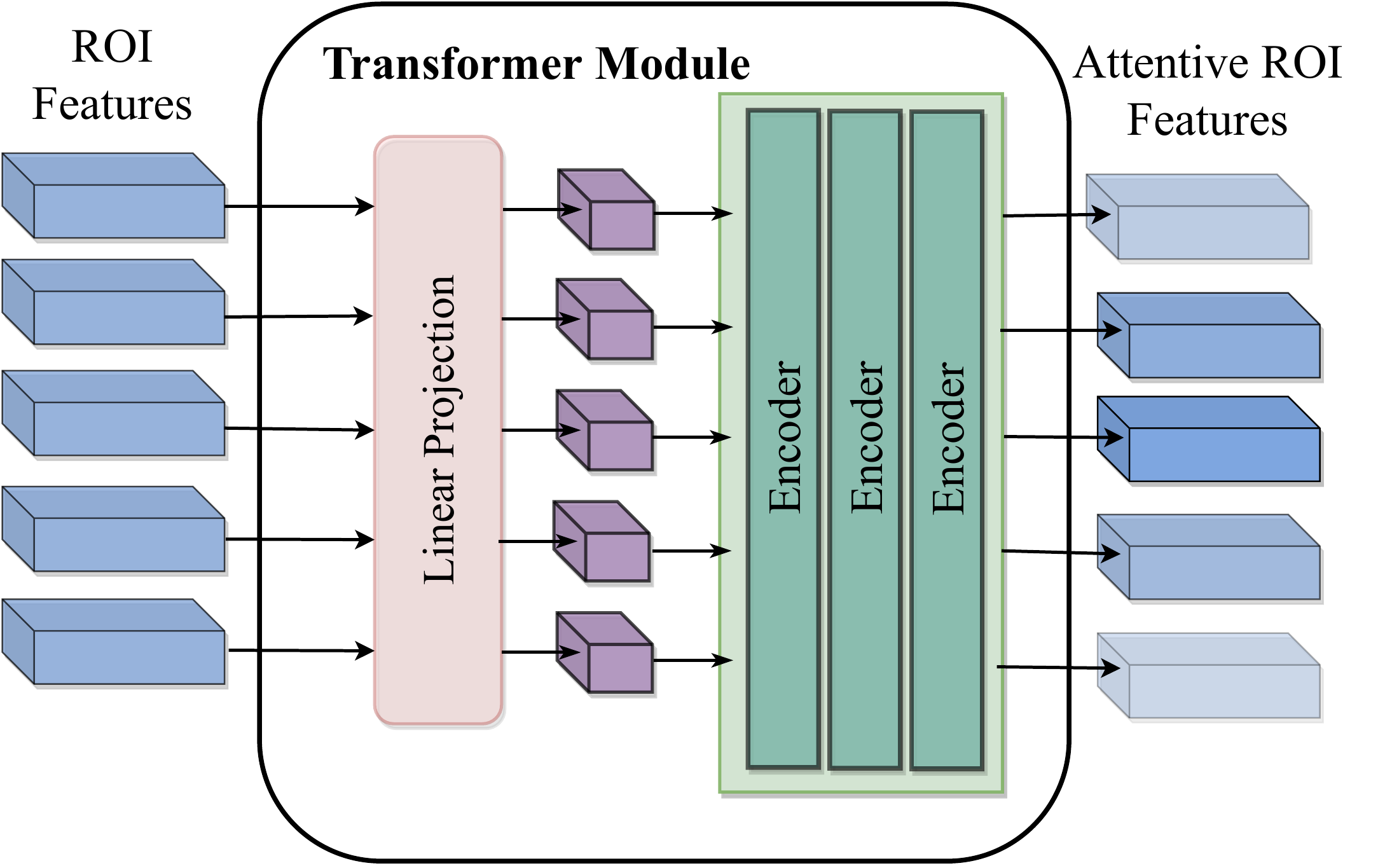}
         \caption{Transformer module}
         \label{fig:left}
     \end{subfigure}
     \hfill
     \begin{subfigure}[b]{0.25\linewidth}

         \includegraphics[width=\linewidth]{ 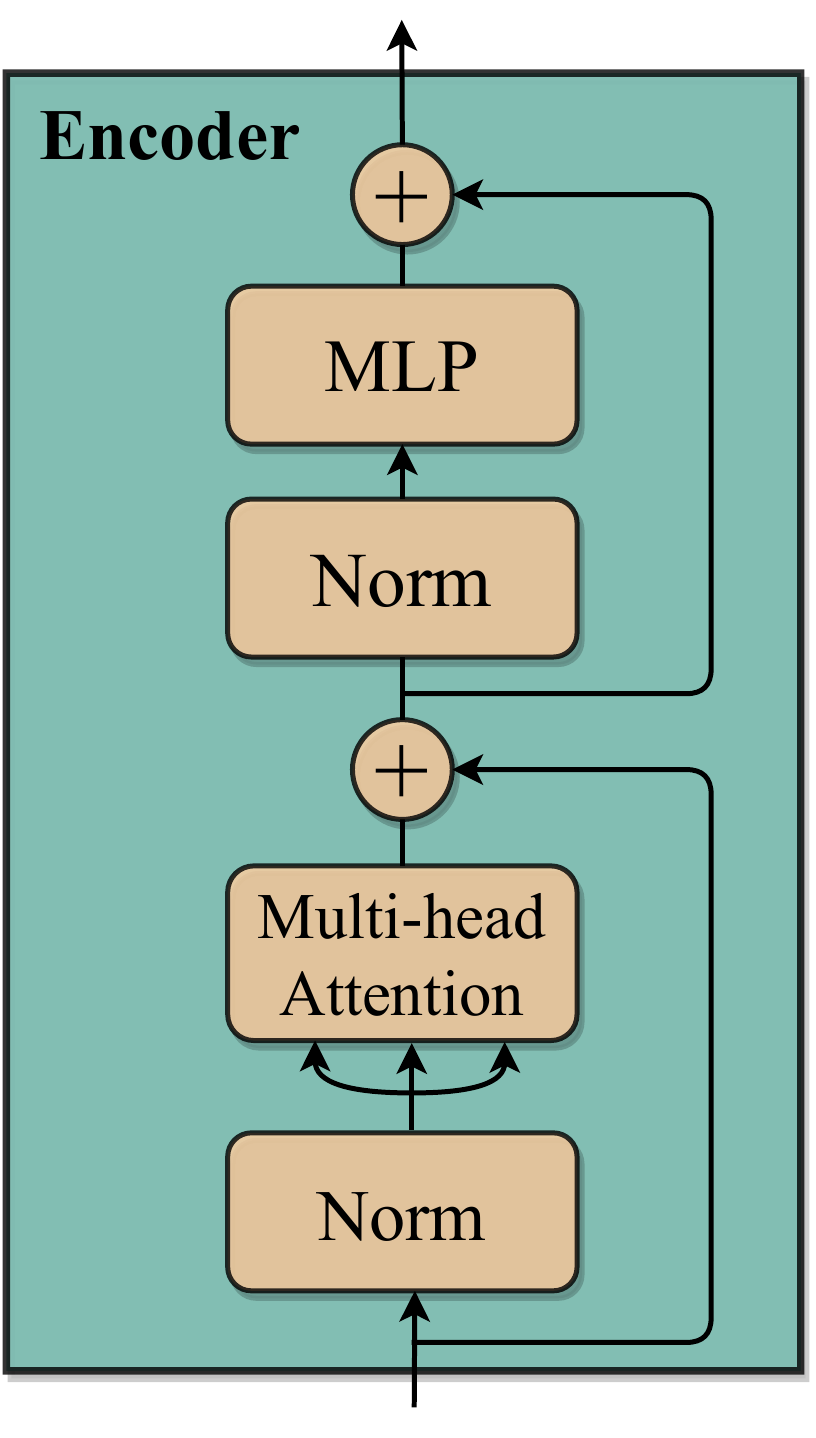}
         \caption{Encoder}
         \label{fig:right}
     \end{subfigure}
     
     \caption{A closer look at the transformer module (a). The region-of-interest (ROI) features are sent as a sequence of tokens to the linear projection layer to form feature embeddings that are in turn input to a set of transformer encoders. (b) This encoder encodes each token as a series of values and corresponding key-query pairs. The resulting sequence is formed by a summation using the learned attention weights that represents the cross-correlation between each element in the input sequence.}
     \vspace{-1.2em}
     
     \label{fig:trans}
 \end{figure}

In the presence of noisy labels, representation learning through feature clustering methods like those in \cite{ZhangProto,lvq,yang2018robust,jiang2018learning} may compute corrupted class prototypes. In object detection, region features of different classes may be similar (such as the ``Car" and ``Truck" categories, see Figure \ref{fig:compare_bb}), rendering outlier removal methods ineffective in cases of mis-classification. In order to address this, we propose a transformer that utilizes self-attention to focus on salient regions-of-interest for prototype computation. A close-up view of the transformer model is shown in Figure \ref{fig:trans}.

Consider the set of features of positive regions-of-interest (ROIs) $R_{feat} = \{f_{i}\}_{i\in N_{roi^+}}$ consisting of feature vectors generated by the object detector $\phi_s$ of ROIs categorized as belonging to the object class. In order to create a representative prototype, we take inspiration from \cite{vit} and send the ROI features as tokens to a transformer module consisting of a linear embedding layer and a set of transformer encoders. The encoder contains alternating multi-head attention blocks and feed-forward blocks, with interspersed normalization layers and residual connections, as depicted in Figure \ref{fig:right}.

The input to the transformer module is the set of positive ROI features associated with the object category. Each feature $f_i$ in $\boldsymbol{f}$ (comparable to the image patches in \cite{vit}) is sent as a token to the linear projection layer,  which gives a set of feature embeddings $\boldsymbol{f'} \in \mathbb{R}^{N_{roi}\times D}$ that are input to the encoder block. The multi-head self attention layer (MSA) \cite{vaswani2017attention}, calls several self-attention operations in parallel in which a linear layer is used to encode and represent each token in $\boldsymbol{f}'$ as a value $\boldsymbol{v}$ and a corresponding query $\boldsymbol{q}_i$ and key $\boldsymbol{k}_j$ pair. A weighted sum of all values in $\boldsymbol{f}'$ is computed for each token where the attention weight is
\setlength{\belowdisplayskip}{0pt} \setlength{\belowdisplayshortskip}{0pt}
\setlength{\abovedisplayskip}{0pt} \setlength{\abovedisplayshortskip}{0pt}
\begin{equation}
    A_{ij} = \text{softmax}\left(\frac{\boldsymbol{q}\boldsymbol{k}^T}{\sqrt{d}}\right),
\end{equation}
where $d$ is a scaling factor. The output of the self attention operation is thus
\begin{equation}
    \text{attention}(\boldsymbol{q},\boldsymbol{k},\boldsymbol{v}) =   A_{ij}\boldsymbol{v}.
\end{equation}
The feed-forward block is a two-layers multi-layer perceptron (MLP) with the GELU \cite{hendrycks2016gaussian} non-linearity function. Following the forward pass through $L$ such encoders, the final output of the transformer module is a set of attentive region features $R^{att}_{feat} = \{f^{att}_{i}\}_{i\in N_{roi^+}}$. By virtue of the self-attention mechanism, we learn the cross-correlation between positive region features, and in turn the ROIs that contribute salient information for prototype computation. 
\noindent\textbf{Predictive entropy.}
As a way to obtain additional insight on how informative each region feature is for the bounding box categorization task, we form the representative attentive class prototype as the sum of the attentive region features $f^{c_j}_{att}$ weighted with the predictive entropy \cite{wang2014new,prabhu2021active} of the classifier $C_{\phi_s}$, denoted by $\mathcal{H}(Y|\boldsymbol{x};\Theta)$. Instead of utilizing the uncertainty of the model as in \cite{prabhu2021active},  we weigh each attentive region feature with the confidence of the associated prediction. The predictive entropy and the resulting entropy weights are given by 
 \begin{equation}
 \nonumber   \mathcal{H}(Y|\boldsymbol{x};\Theta) = -\sum_{c=1}^{N_c}p_\Theta(Y=c|\boldsymbol{x})\log\{p_\Theta(Y=c|\boldsymbol{x})\},
\end{equation}
\begin{equation}
    \boldsymbol{E} = 1 - \frac{\mathcal{H}(Y|\boldsymbol{x};\Theta)}{\sum_{x\in X}\mathcal{H}(Y|\boldsymbol{x};\Theta)}.
\end{equation}
The attentive prototype as the weighted average of the attentive ROI features is obtained as follows
\begin{equation}
    F^{att} = \frac{1}{N^{roi^+}}\sum_{i=1}^{N_{roi^+}}E_i f^{att}_i.
\end{equation}

\noindent\textbf{Computing the final prototype.} During the process of training the object detector for adaptation, samples from the target dataset $X^t=\{x^T_j\}_{j=1}^M$ are sent in mini-batches. Each attentive prototype computed in an iteration is combined with the prototoype computed in the previous iteration through exponential moving average (EMA). The initial prototype at the first iteration of the first epoch is simply the average of the positive ROI features. The final attentive prototype at each iteration $j$ is thus given by
 \begin{equation}
    F^{final}_{j} = \alpha F^{final}_{j-1} + (1-\alpha) F^{attentive}_{j},
\end{equation}
with a keep ratio $\alpha=0.9999$ used in our experiments. 
\subsection{Similarity-based refinement}
Once the representative class prototype is obtained, we use it as a soft-filter to identify region features that are dissimilar and thus far away from each other in the feature space. To do this, the distance of each positive ROI in $R_{feat}$ from its corresponding prototype is calculated using the metric of cosine similarity such that
\begin{equation}
    d^i = \frac{\sum_{k=1}^{K}F^{final}_{k} \cdot f^{k}_i}{\sqrt{\sum_{k=1}^{K}(F^{final}_{k})^2}\sqrt{\sum_{k=1}^{K}(f^{k}_i)^2}},
\end{equation}
where $K$ is the feature dimension and $i$ ranges from $1$ to the number of ROIs, $N_{roi}$. Cosine similarity is chose due to the sparse nature of the features. The classification loss corresponding to each positive region of interest is multiplied by a similarity weight, computed by taking the softmax of the cosine distance, such that
\begin{equation}
    \mathcal{L}'_{cls} = \frac{1}{N_{roi}} (\sum_{i \in roi^+} d^i \ell^i_{cls} + \sum_{j \in roi^-} \ell^j_{cls}), 
\end{equation}

 where $\ell^{i/j}_{cls}$ corresponds to the region-wise loss of the bounding box classifier, where $i$ indexes the positive regions and $j$ indexes the negative regions. With this similarity-based down-weighing, the losses corresponding to regions that have been identified as incorrect through prototype matching will be down-weighed and not contribute to training. As the representative prototype improves with each epoch, the network becomes better at soft-filtering incorrect regions and avoids reinforcing the error in pseudo-labels.

  \begin{algorithm}
\caption{Attentive prototype computation and label refinement.}\label{alg:cap}
\begin{algorithmic}
\Require Source trained model $\phi^s$, unannotated target data $X^T$, pseudo labels $Y^{Ps}$
\Ensure Bounding box predictions
\State - Initialize model $\phi \leftarrow \phi^s$
    \State - $\alpha=0.9999$
      \For{$1\leq it_{m} \leq N_{meta}$}
      \State - Initialize prototype $F^{init}\leftarrow None$
      \For{$1 \leq it \leq N_{it}$}
      \State - $F^{init}\leftarrow F_{epoch-1}$ if $epoch>0$
      \State - Get $R_{feat} = \{f_{i}\}_{i\in N_{roi^+}}$
      \State - $R_{feat}^{att} = \mathcal{T}(R_{feat})$   \algorithmiccomment{send features to transformer} $\mathcal{T}$
      \State - Get classifier o/p  $C_{pred}$
      \State - $P_{pred}= sigmoid(C_{pred})$
      \State -  $\boldsymbol{E}=-(P_{pred})\log(P_{pred}) + (1 - P_{pred})\log(1 - P_{pred})$      \algorithmiccomment{entropy}
      \State - $W_{entropy} = 1- \boldsymbol{E} $
      \State - $F^{att} =  \sum_{i=1}^{N_{roi^+}} W_{entropy} \cdot f_{i}^{attentive} $
      
      \State - $F^{att} \leftarrow \alpha F^{init} + (1-\alpha)F^{att} $
      
      \State -  $d^i = cosine\_similarity(F^{att},f_{i})$
      
      \State - $ \mathcal{L}'_{cls} = \frac{1}{N_{roi}} (\sum_{i \in roi^+} d^i \ell^i_{cls} + \sum_{j \in roi^-} \ell^j_{cls})$

      \State - $F^{init} \leftarrow F^{attentive}$

     \EndFor

     \State - Output predictions $Y^{pred}$
     \State - $Y^{Ps} \leftarrow threshold(Y^{pred})$
     \State - $\phi \leftarrow \phi^{ps}$
     
     \EndFor
\end{algorithmic}
\end{algorithm}

\section{Experiments}
We demonstrate our domain adaptation framework on two base object detection networks, SECOND-iou \cite{yang2021st3d}, which is a modified version of the voxel-based network SECOND \cite{second}, and PointRCNN \cite{shi2019pointrcnn}, a two stage point-based network. We explore three cross-dataset domain shift scenarios for each detector for the ``Car" object category. In this section, we explain the details of the experiments and the datasets used.
\begin{table*}[]
\caption{A tabular comparison of 3D mean average precision (mAP) results of the ``Car" object for the adaptation of two object detection networks SECOND-iou \cite{second}, and PointRCNN \cite{shi2019pointrcnn} against recent domain adaptation methods. Where the target dataset is KITTI, we evaluate with the official metric across 3 difficulty categories for an IoU threshold of $0.7$. In the case of the nuScenes target dataset, we average across the various difficulty categories of the official metric. The best results are in bold type.}
\vspace{-1em}
\label{tab:main-table}
\begin{tabular}{ccccc|ccccc}
\midrule[1pt]\toprule[0.1pt]
\multicolumn{5}{c|}{SECOND-iou}                                                                                                                           & \multicolumn{5}{c}{Point-RCNN}                                                                                                                                         \\ \hline
\multicolumn{1}{c|}{\multirow{2}{*}{Domain shift}}      & \multicolumn{1}{c|}{\multirow{2}{*}{Method}} & \multicolumn{3}{c|}{mAP}                         & \multicolumn{1}{c|}{\multirow{2}{*}{Domain shift}}      & \multicolumn{1}{c|}{\multirow{2}{*}{Method}} & \multicolumn{3}{c}{mAP}                                       \\ \cline{3-5} \cline{8-10} 
\multicolumn{1}{c|}{}                                   & \multicolumn{1}{c|}{}                        & easy           & mod.           & hard           & \multicolumn{1}{c|}{}                                   & \multicolumn{1}{c|}{}                        & easy  & mod.                      & hard                      \\ \midrule[0.1pt]\toprule[0.1pt]
\multicolumn{1}{c|}{\multirow{6}{*}{Waymo $\rightarrow$ KITTI}}    & \multicolumn{1}{c|}{DT}                      & 46.66          & 39.86          & 35.03          & \multicolumn{1}{c|}{\multirow{6}{*}{Waymo $\rightarrow$ KITTI}}    & \multicolumn{1}{c|}{DT}                      & 13.11 & \multicolumn{1}{l}{12.10} & \multicolumn{1}{l}{12.24} \\
\multicolumn{1}{c|}{}                                   & \multicolumn{1}{c|}{ST}                      & 69.04          & 55.77          & 54.57          & \multicolumn{1}{c|}{}                                   & \multicolumn{1}{c|}{ST}                      & 21.13 & 19.29 & 18.28             \\
\multicolumn{1}{c|}{}                                   & \multicolumn{1}{c|}{SN\cite{Wang2020TrainIG}}                      & 73.23          & 56.36          & 54.25          & \multicolumn{1}{c|}{}                                   & \multicolumn{1}{c|}{SN}                      & 48.7 & 47.1                    & \textbf{49.7}                     \\
\multicolumn{1}{c|}{}                                   & \multicolumn{1}{c|}{ST3D\cite{yang2021st3d}}                    & 67.47          & 59.17          & 56.73          & \multicolumn{1}{c|}{}                                   & \multicolumn{1}{c|}{SF-UDA\textsuperscript{3D}\cite{Saltori2020SFUDA3DSU}}                   & -     & -                         & -                         \\
\multicolumn{1}{c|}{}                                   & \multicolumn{1}{c|}{Proposed}                & \textbf{75.75}     & \textbf{64.43}      & \textbf{57.19}   & \multicolumn{1}{c|}{}                                   & \multicolumn{1}{c|}{Proposed}                &  \textbf{62.11}    &  \textbf{53.08}     &  46.64   \\ \cline{2-5} \cline{7-10} 
\multicolumn{1}{c|}{}                                   & \multicolumn{1}{c|}{Oracle}                  & 84.86          & 68.93          & 67.38          & \multicolumn{1}{c|}{}                                   & \multicolumn{1}{c|}{Oracle}                  & 81.61 & 74.36                     & 74.49                     \\ \midrule[0.1pt]\toprule[0.1pt]
\multicolumn{1}{c|}{\multirow{6}{*}{nuScenes $\rightarrow$ KITTI}} & \multicolumn{1}{c|}{DT}                      & 18.37          & 17.31          & 16.09          & \multicolumn{1}{c|}{\multirow{6}{*}{nuScenes $\rightarrow$ KITTI}} & \multicolumn{1}{c|}{DT}                      & 10.59 & 10.76                     & 10.64                     \\
\multicolumn{1}{c|}{}                                   & \multicolumn{1}{c|}{ST}                      & 53.03          & 37.71          & 35.18          & \multicolumn{1}{c|}{}                                   & \multicolumn{1}{c|}{ST}                      & 22.21 & 11.56                     & 11.90                     \\
\multicolumn{1}{c|}{}                                   & \multicolumn{1}{c|}{SN}                      & 22.03          & 18.51          & 18.04          & \multicolumn{1}{c|}{}                                   & \multicolumn{1}{c|}{SN}                      & 60.35 & 54.82                     & 52.78                     \\
\multicolumn{1}{c|}{}                                   & \multicolumn{1}{c|}{ST3D}                    & 58.24          & 43.13          & 39.46          & \multicolumn{1}{c|}{}                                   & \multicolumn{1}{c|}{SF-UDA\textsuperscript{3d}}                   & 68.8  & 49.80                     & 45.0                      \\
\multicolumn{1}{c|}{}                                   & \multicolumn{1}{c|}{Proposed}                & \textbf{71.56} & \textbf{52.12} & \textbf{45.86} & \multicolumn{1}{c|}{}                                   & \multicolumn{1}{c|}{Proposed}                &  \textbf{69.98}    &\textbf{61.43}& \multicolumn{1}{l}{\textbf{54.26}}    \\ \cline{2-5} \cline{7-10} 
\multicolumn{1}{c|}{}                                   & \multicolumn{1}{c|}{Oracle}                  & 84.86          & 68.93          & 67.38          & \multicolumn{1}{c|}{}                                   & \multicolumn{1}{c|}{Oracle}                                      & 81.61 & 74.36                     & 74.49                     \\ \midrule[0.1pt]\toprule[1pt]
\end{tabular}
\end{table*}
\begin{table*}[]
\centering
\vspace{-1em}
\begin{tabular}{ccc|ccc}
\midrule[1pt]\toprule[0.1pt]
\multicolumn{3}{c|}{SECOND-iou}                                                                          & \multicolumn{3}{c}{Point-RCNN}                                                                  \\ \hline
\multicolumn{1}{c|}{Domain shift}                       & \multicolumn{1}{c|}{Method}   & mAP            & \multicolumn{1}{c|}{Domain shift}                       & \multicolumn{1}{c|}{Method}   & mAP   \\ \hline
\multicolumn{1}{c|}{\multirow{6}{*}{Waymo $\rightarrow$ nuScenes}} & \multicolumn{1}{c|}{DT}       & 15.25          & \multicolumn{1}{c|}{\multirow{6}{*}{KITTI $\rightarrow$ nuScenes}} & \multicolumn{1}{c|}{DT}       & 10.07 \\
\multicolumn{1}{c|}{}                                   & \multicolumn{1}{c|}{ST}       & 17.80          & \multicolumn{1}{c|}{}                                   & \multicolumn{1}{c|}{ST}       & 16.78 \\
\multicolumn{1}{c|}{}                                   & \multicolumn{1}{c|}{SN}       & 18.57          & \multicolumn{1}{c|}{}                                   & \multicolumn{1}{c|}{SN}       & 18.7  \\
\multicolumn{1}{c|}{}                                   & \multicolumn{1}{c|}{ST3D}     & 20.19          & \multicolumn{1}{c|}{}                                   & \multicolumn{1}{c|}{SF-UDA\textsuperscript{3d}}    & \textbf{26.8}  \\
\multicolumn{1}{c|}{}                                   & \multicolumn{1}{c|}{Proposed} & \textbf{20.47} & \multicolumn{1}{c|}{}                                   & \multicolumn{1}{c|}{Proposed} &     18.87  \\ \cline{2-3} \cline{5-6} 
\multicolumn{1}{c|}{}                                   & \multicolumn{1}{c|}{Oracle}   & 32.64          & \multicolumn{1}{c|}{}                                   & \multicolumn{1}{c|}{Oracle}   & 29.89 \\ \midrule[0.1pt]\toprule[1pt]
\end{tabular}
\vspace{-2em}
\end{table*}

\subsection{Experimental setup}

\noindent\textbf{Datasets.} In order to simulate the various domain shifts, we consider three popular large-scale autonomous driving datasets with considerable domain gaps among them, The Waymo Open Dataset \cite{waymo}, the KITTI dataset \cite{KITTI}, and the nuScenes dataset \cite{nuscenes2019}. The largest dataset among these is Waymo, with more than 230K annotated lidar frames collected across six US cities, of which we use approximately 50K due to memory constraints in a 40K/10K training/validation (test) split. The nuScenes dataset consists of 34,149 frames which we utilize in 28K/6K split. The smallest dataset is KITTI, consisting of 7,481 (3K/3K split) annotated lidar frames collected from Germany. All training and validation splits used are official and consistent with that used by other works in the 3D object detection literature.

\noindent\textbf{Object detection networks.}
The network SECOND-iou \cite{second,yang2021st3d} is a two stage voxel-based 3D object detector which uses a PointNet \cite{REF:qi2017pointnet} backbone to extract voxel features from groupings of points and consists of a grouping layer, a region proposal network (RPN) and an ROI refinement head. It is a modified version of SECOND \cite{second} proposed in \cite{yang2021st3d}, with the extra refinement head. The region features for prototype computation are obtained at the output of the RPN head, and the classification loss at this stage is down-weighed during adaptation. PointRCNN is a two stage point-based network with a similar PointNet backbone for feature extraction that generates 3D region proposals through a bottom-up approach through foreground segmentation. This is followed by a refinement network. We implement adaptation for PointRCNN in this second stage.

\subsection{Implementation details}
In addition to the mentioned refinement steps, we implement an iterative training strategy in which the domain adaptive network is trained and re-trained a series of times with the source model $\phi_s$ and pseudo labels re-initialized to the previously trained model and generated predictions respectively. For the implementation of SECOND-iou, we use the public pyTorch repository Open-PCDet \cite{openpcdet2020}, and the official code release from \cite{shi2019pointrcnn} for the implementation of PointRCNN. We perform experiments with a 48GB Quadro RTX 8000 GPU and a 16GB GeForce RTX 2080 GPU. We follow the lengths of training recommended by the authors in the case of each object detector network. In both cases, the cyclic Adam optimization algorithm is used.

\section{Results}
In this section we demonstrate the results of our domain adaptation framework and compare it against four domain adaptation methods;  \textit{Direct transfer (DT)}: Inference of the source trained model on target data, \textit{Current SOTA}: ST3D \cite{yang2021st3d} by Yang \etal and SFUDA\textsuperscript{3D} \cite{Saltori2020SFUDA3DSU}  by Saltori \etal for their corresponding base networks, \textit{Statistical normalization (SN) \cite{Wang2020TrainIG}}: weakly supervised approach that uses the target domain bounding-box statistics, \textit{Pseudo-label self training (ST)}: Re-training the object detector on thresholded source-model generated pseudo-labels. For reference we, also compare with the ``oracle" results, which are obtained by training the detector with the ground truth labels of the target domain, indicating the possible upper bound of performance after adaptation.

\begin{figure*}[ht]
  \centering
\begin{tabular}[!t]{ccccccc}

& & \hspace{-12em}\textbf{SECOND-iou \cite{second}}& &

\\

  \includegraphics[trim={0cm 0cm 0cm 0cm},clip,width=0.245\linewidth]{ 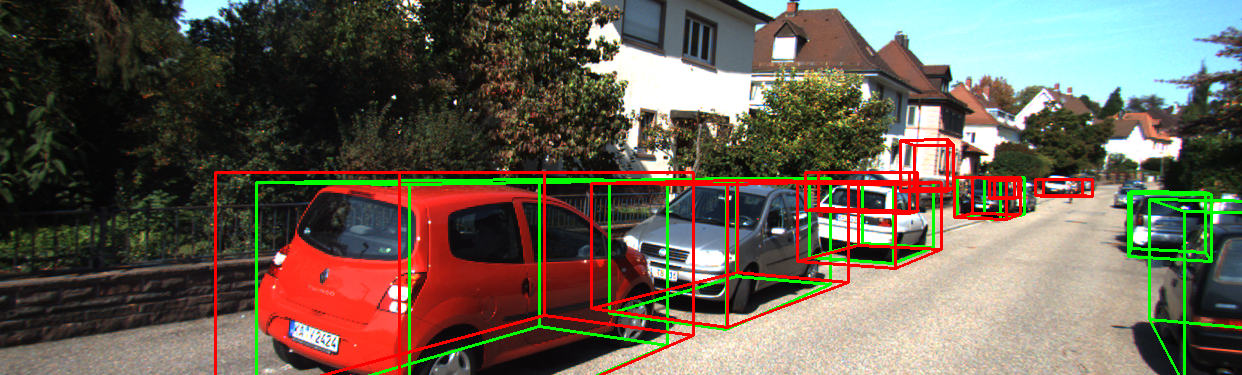} \hspace{-1.2em} &  
  \includegraphics[trim={0cm 0cm 0cm 0cm},clip,width=0.245\linewidth]{ 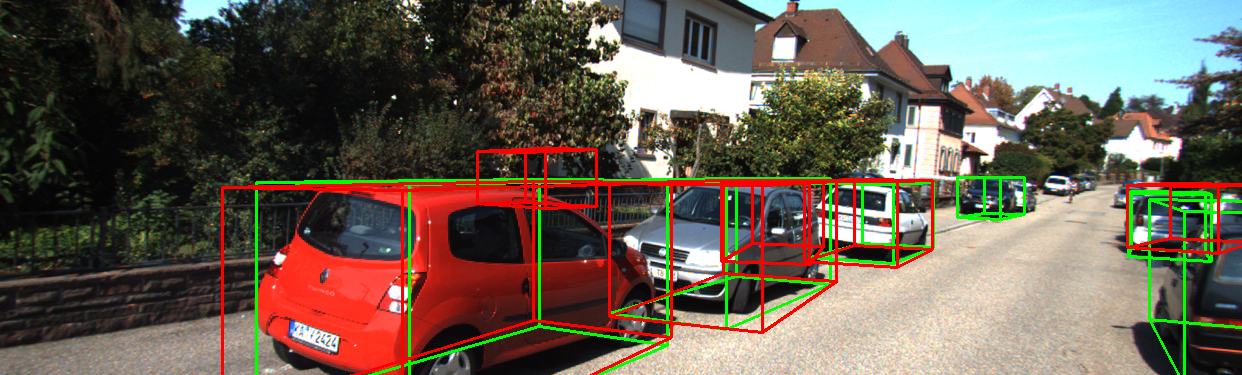} \hspace{-1.2em} & 
  \includegraphics[trim={0cm 0cm 0cm 0cm},clip,width=0.245\linewidth]{ 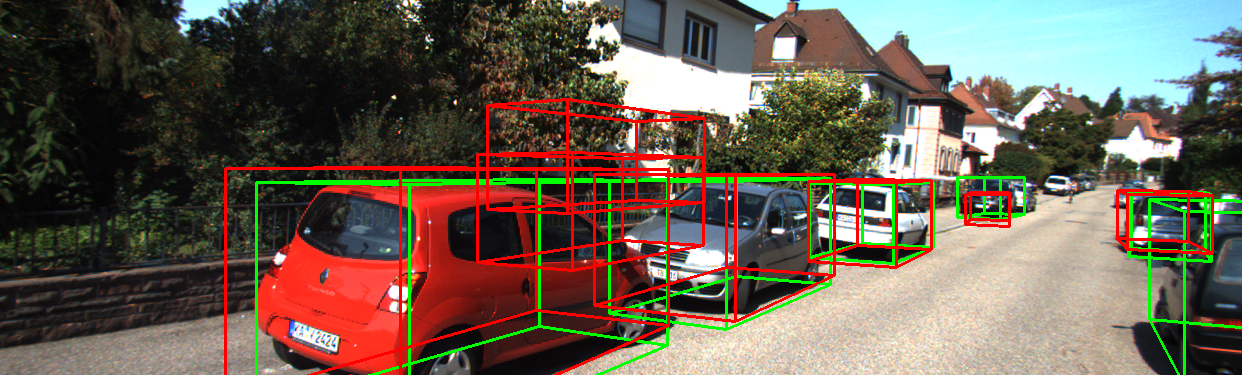} \hspace{-1.2em} & 
  \includegraphics[trim={0cm 0cm 0cm 0cm},clip,width=0.245\linewidth]{ 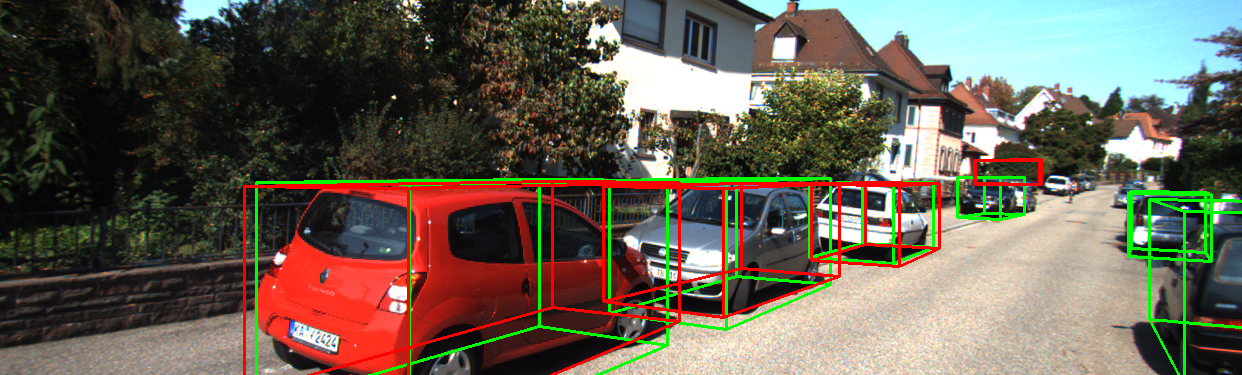} \hspace{-1.2em} & 
  
  \\

    \includegraphics[trim={0cm 0cm 0cm 0cm},clip,width=0.245\linewidth]{ 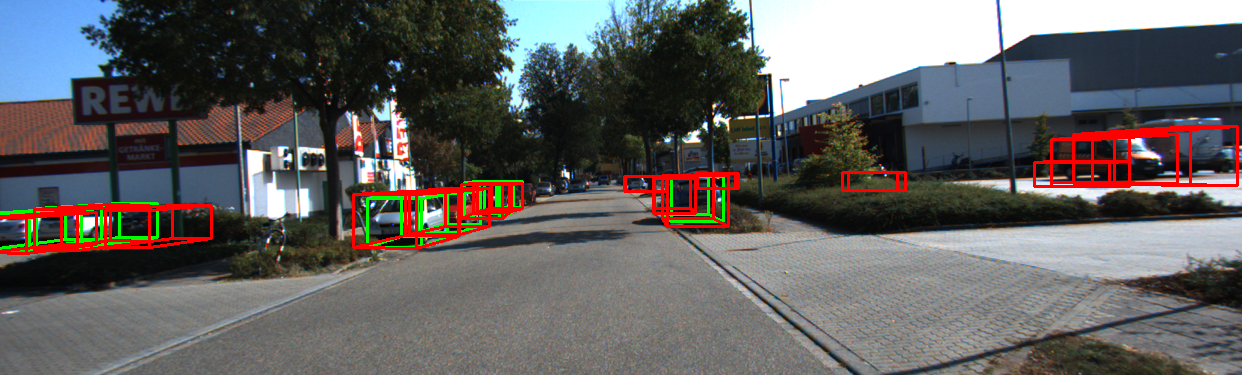} \hspace{-1.2em} &  
  \includegraphics[trim={0cm 0cm 0cm 0cm},clip,width=0.245\linewidth]{ 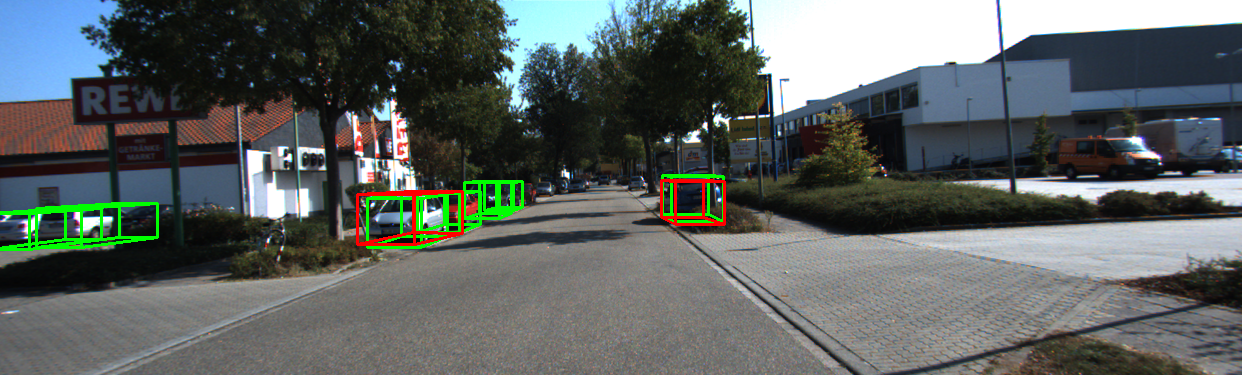} \hspace{-1.2em} & 
  \includegraphics[trim={0cm 0cm 0cm 0cm},clip,width=0.245\linewidth]{ 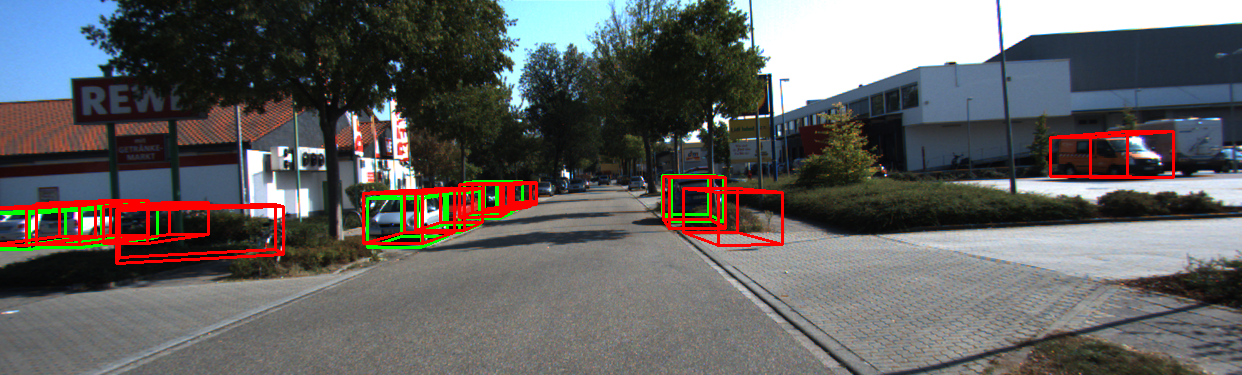} \hspace{-1.2em} & 
  \includegraphics[trim={0cm 0cm 0cm 0cm},clip,width=0.245\linewidth]{ 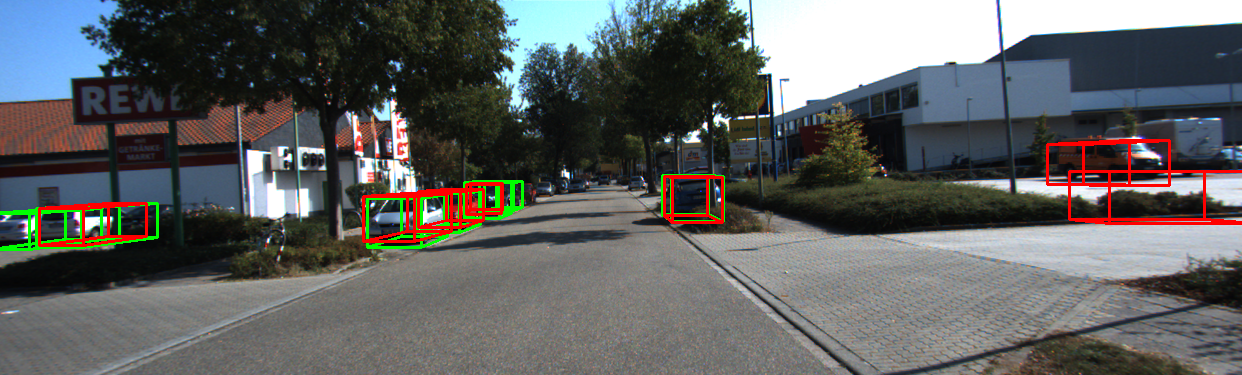} \hspace{-1.2em} & 
  
  \\

      \includegraphics[trim={0cm 0cm 0cm 0cm},clip,width=0.245\linewidth]{ 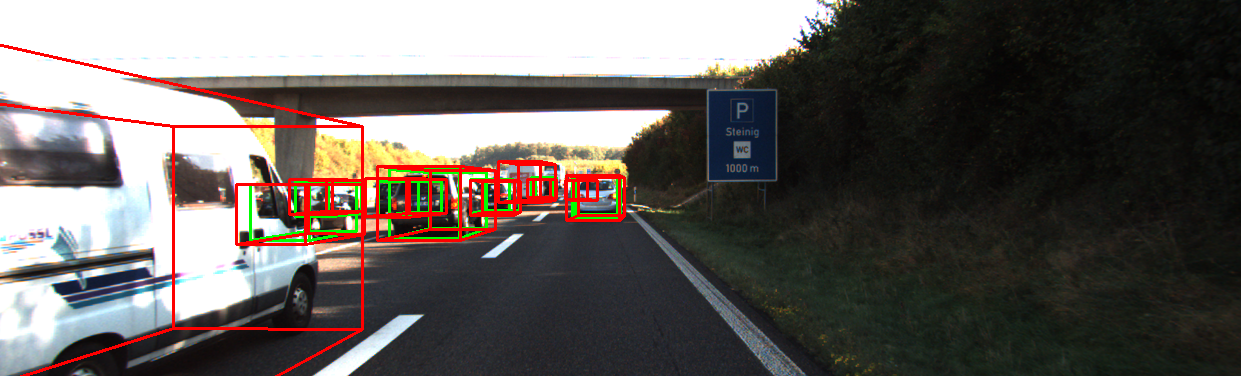} \hspace{-1.2em} &  
  \includegraphics[trim={0cm 0cm 0cm 0cm},clip,width=0.245\linewidth]{ 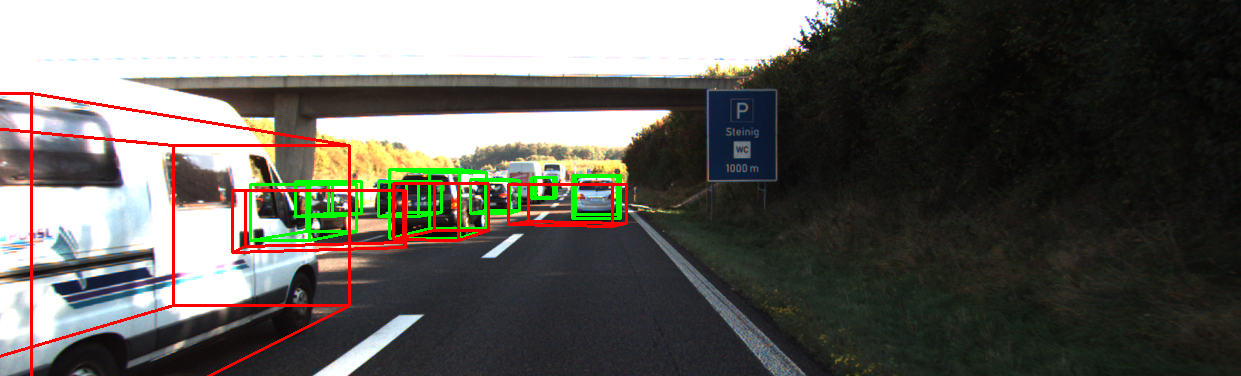} \hspace{-1.2em} & 
  \includegraphics[trim={0cm 0cm 0cm 0cm},clip,width=0.245\linewidth]{ 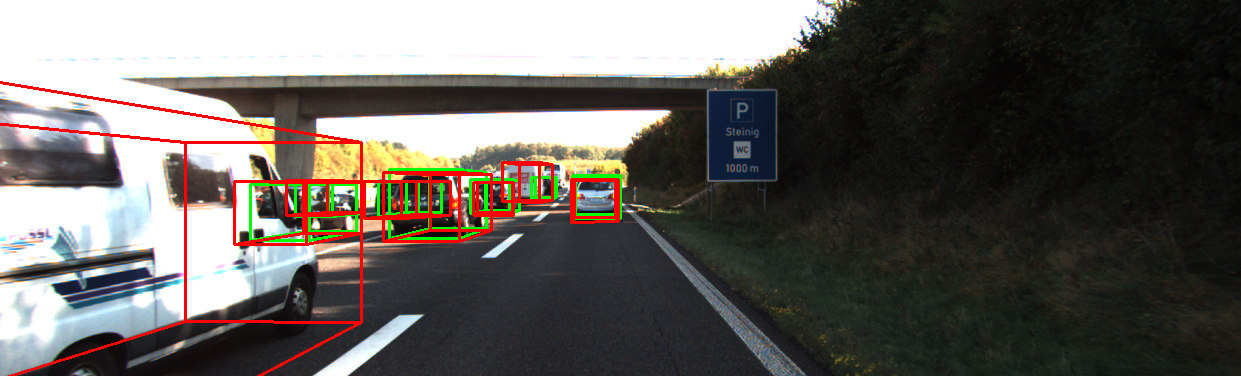} \hspace{-1.2em} & 
  \includegraphics[trim={0cm 0cm 0cm 0cm},clip,width=0.245\linewidth]{ 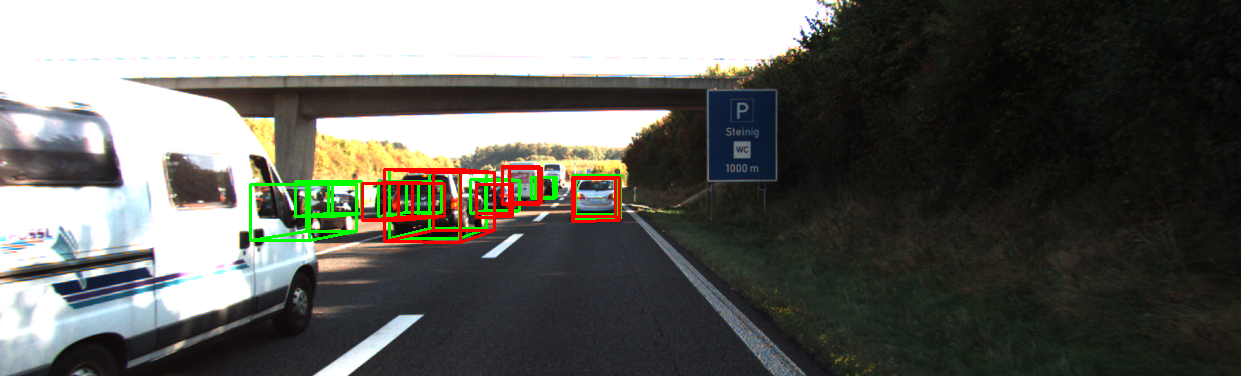} \hspace{-1.2em} & 
  
  \\

      \includegraphics[trim={0cm 0cm 0cm 0cm},clip,width=0.245\linewidth]{ 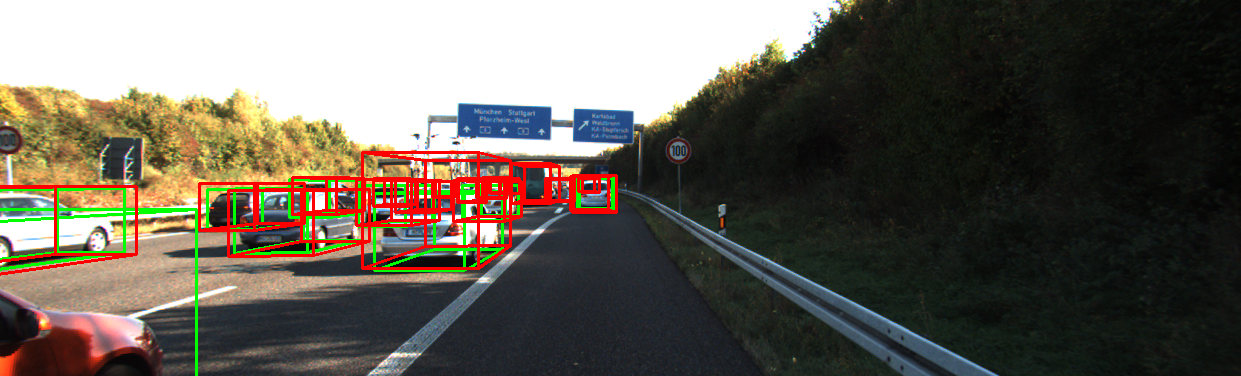} \hspace{-1.2em} &  
  \includegraphics[trim={0cm 0cm 0cm 0cm},clip,width=0.245\linewidth]{ 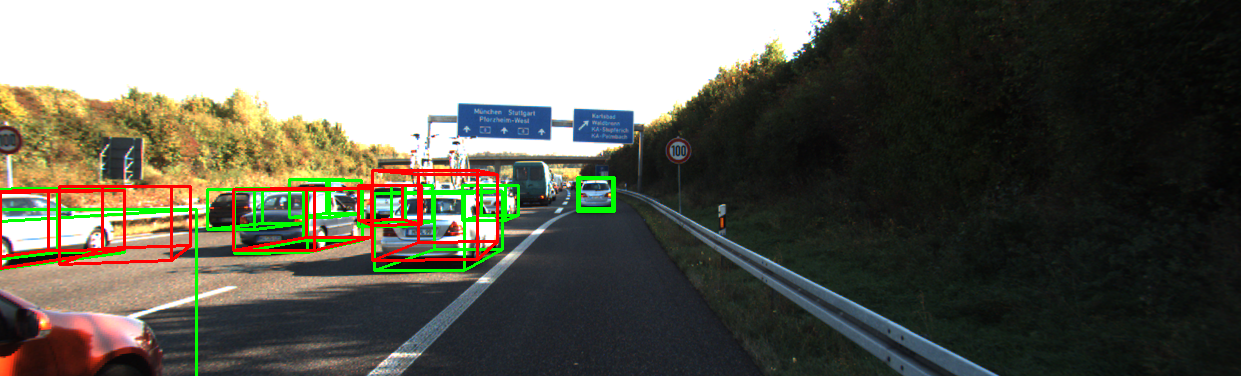} \hspace{-1.2em} & 
  \includegraphics[trim={0cm 0cm 0cm 0cm},clip,width=0.245\linewidth]{ 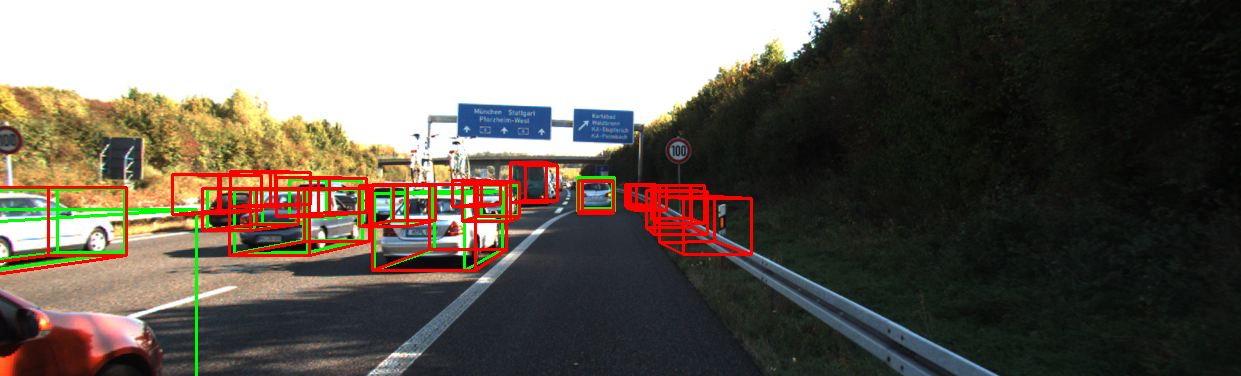} \hspace{-1.2em} & 
  \includegraphics[trim={0cm 0cm 0cm 0cm},clip,width=0.245\linewidth]{ 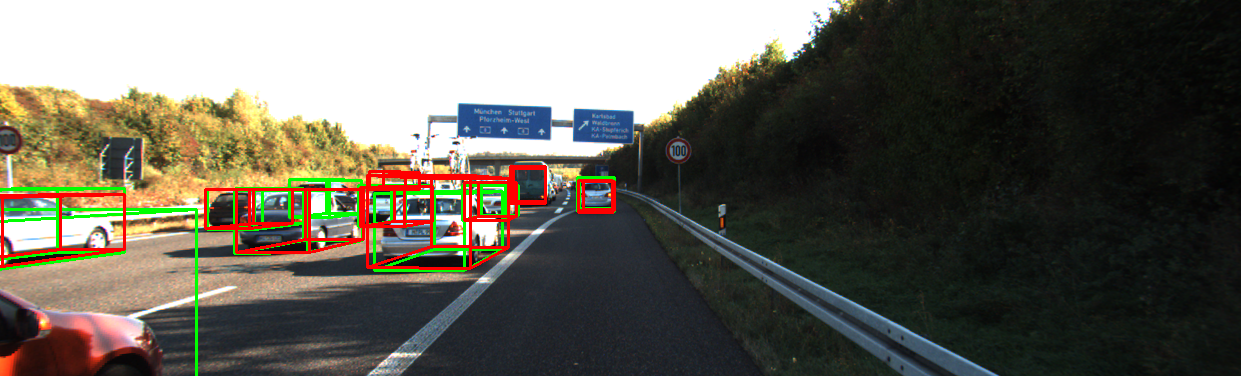} \hspace{-1.2em} & 
  
  \\
    
  Direct transfer \hspace{-1.2em} & ST \hspace{-1.2em}  & ST3D \cite{yang2021st3d} \hspace{-1.2em} & Ours \hspace{-1.2em}
  
  \\
  
  & & \hspace{-12em}\textbf{PointRCNN \cite{shi2019pointrcnn}}& &
  
  \\

    \includegraphics[trim={0cm 0cm 0cm 0cm},clip,width=0.245\linewidth]{ 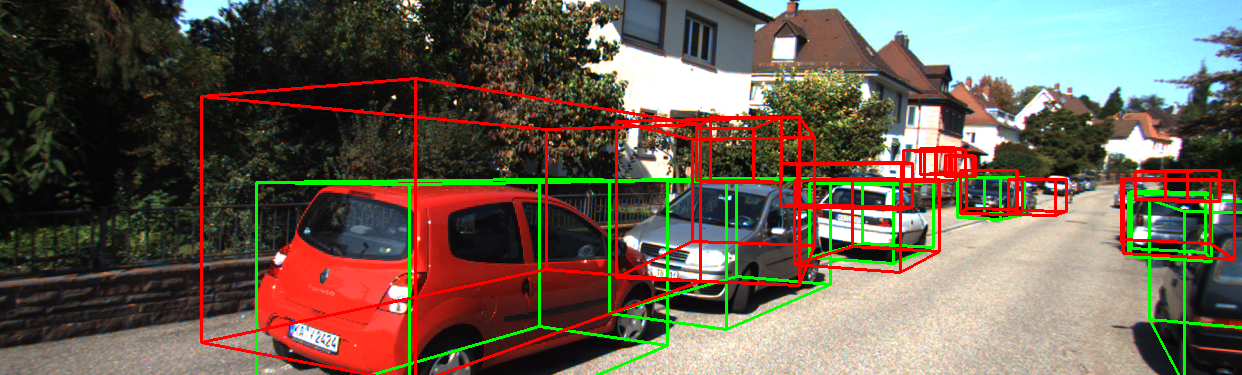} \hspace{-1.2em} &  
  \includegraphics[trim={0cm 0cm 0cm 0cm},clip,width=0.245\linewidth]{ 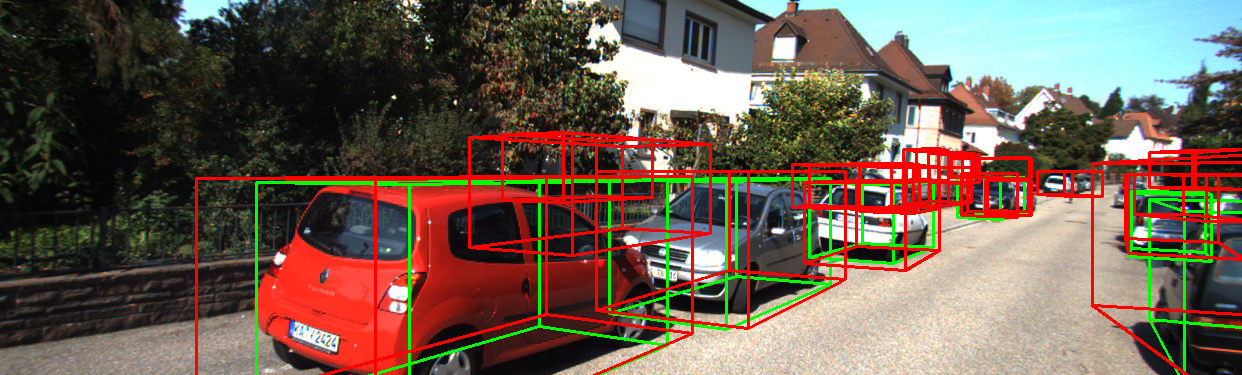} \hspace{-1.2em} & 
  \includegraphics[trim={0cm 0cm 0cm 0cm},clip,width=0.245\linewidth]{ 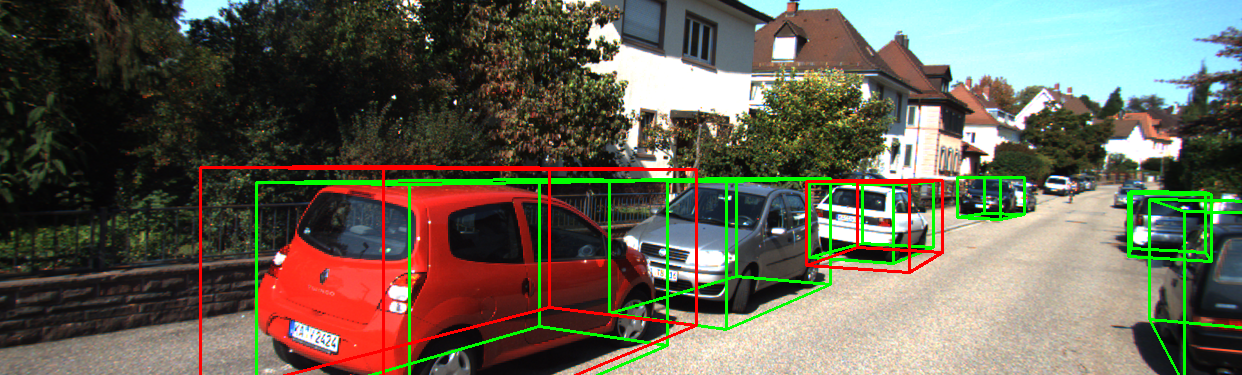} \hspace{-1.2em} & 
  \includegraphics[trim={0cm 0cm 0cm 0cm},clip,width=0.245\linewidth]{ 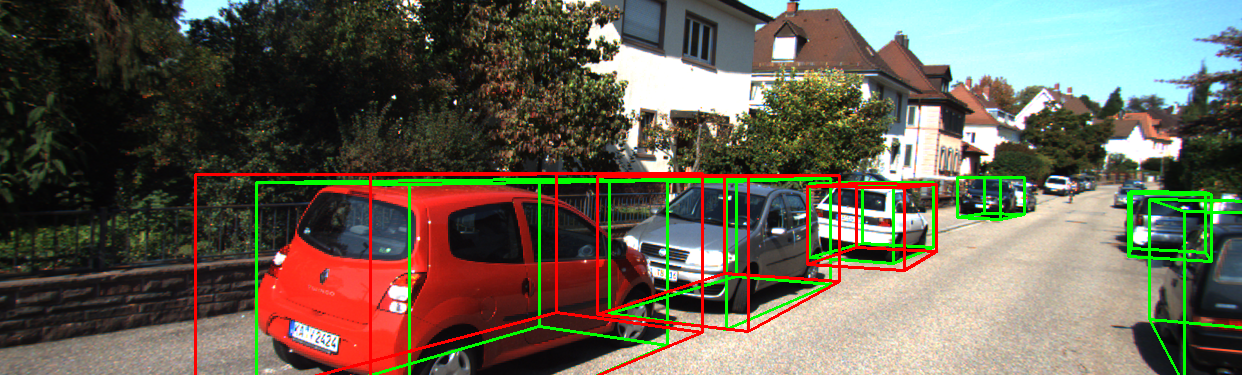} \hspace{-1.2em} & 
  
  \\

    \includegraphics[trim={0cm 0cm 0cm 0cm},clip,width=0.245\linewidth]{ 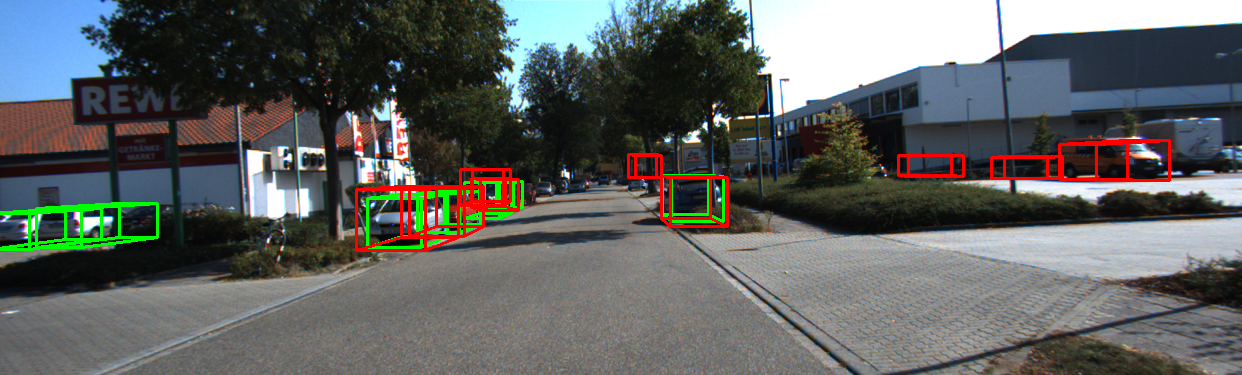} \hspace{-1.2em} &  
  \includegraphics[trim={0cm 0cm 0cm 0cm},clip,width=0.245\linewidth]{ 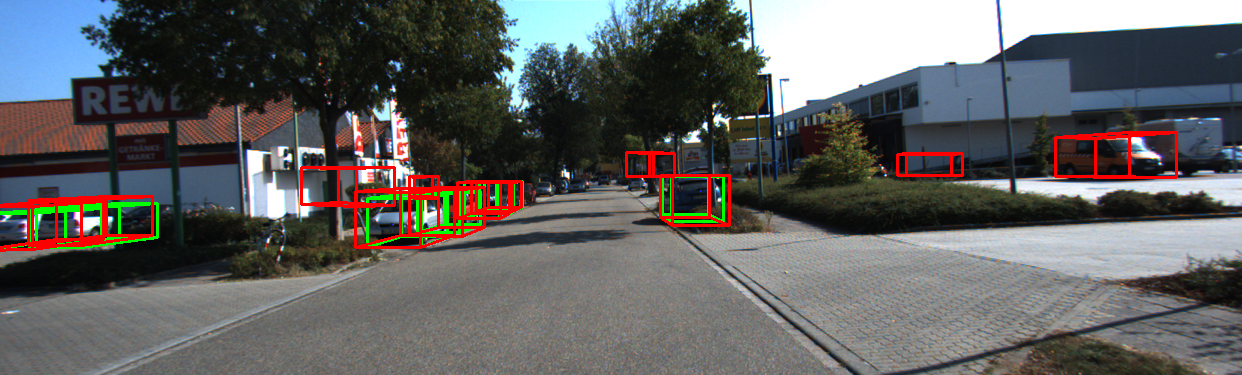} \hspace{-1.2em} & 
  \includegraphics[trim={0cm 0cm 0cm 0cm},clip,width=0.245\linewidth]{ 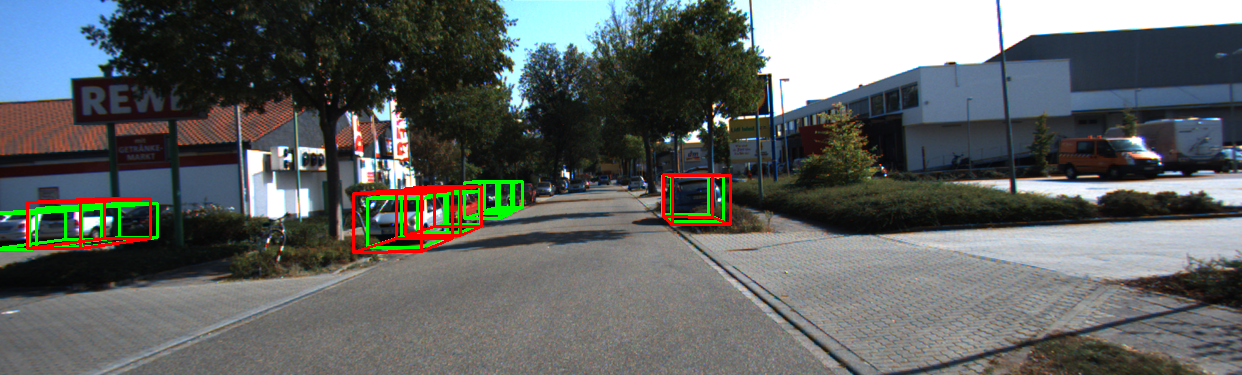} \hspace{-1.2em} & 
  \includegraphics[trim={0cm 0cm 0cm 0cm},clip,width=0.245\linewidth]{ 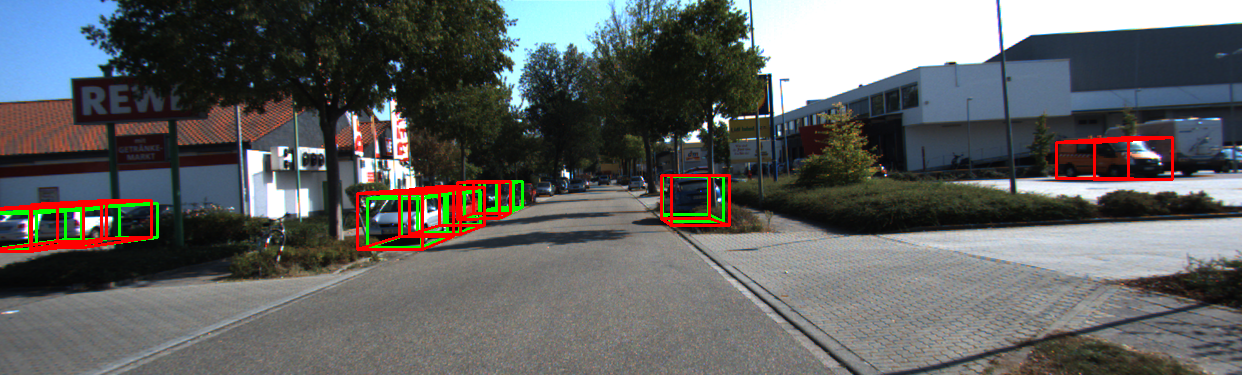} \hspace{-1.2em} & 
  
  \\

      \includegraphics[trim={0cm 0cm 0cm 0cm},clip,width=0.245\linewidth]{ 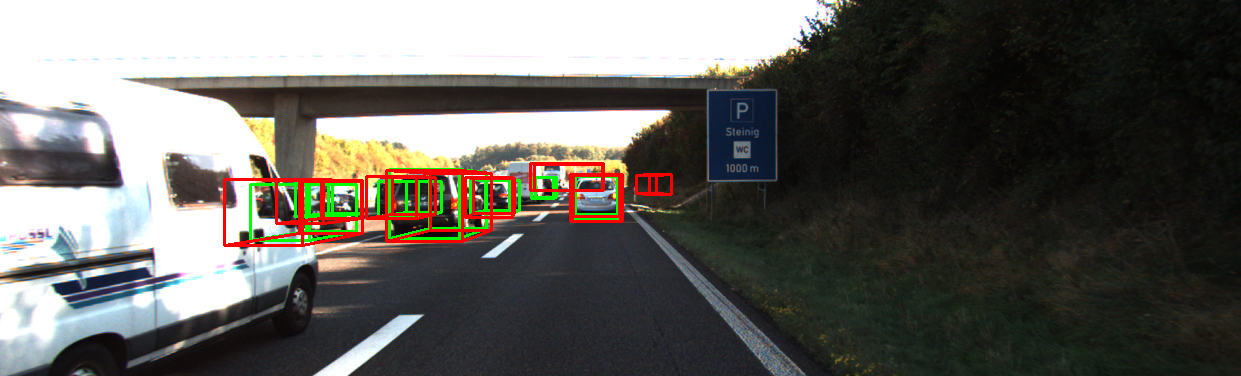} \hspace{-1.2em} &  
  \includegraphics[trim={0cm 0cm 0cm 0cm},clip,width=0.245\linewidth]{ 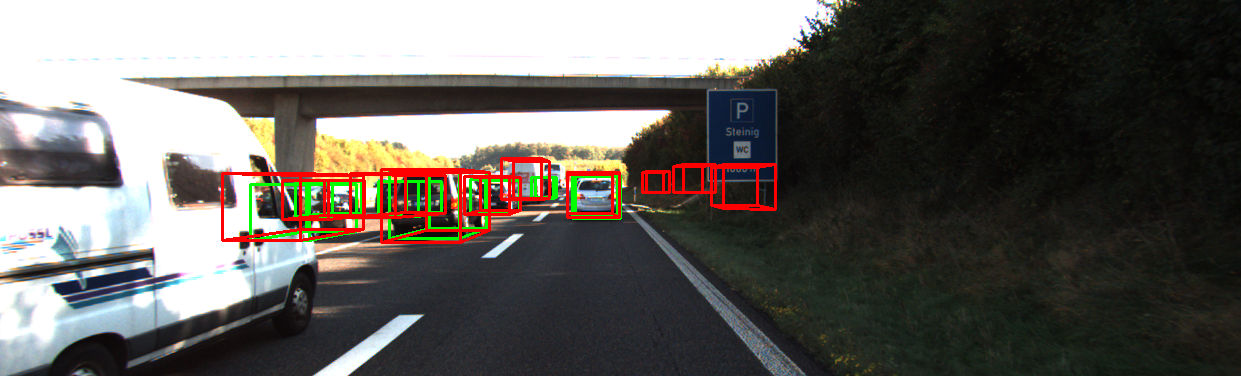} \hspace{-1.2em} & 
  \includegraphics[trim={0cm 0cm 0cm 0cm},clip,width=0.245\linewidth]{ 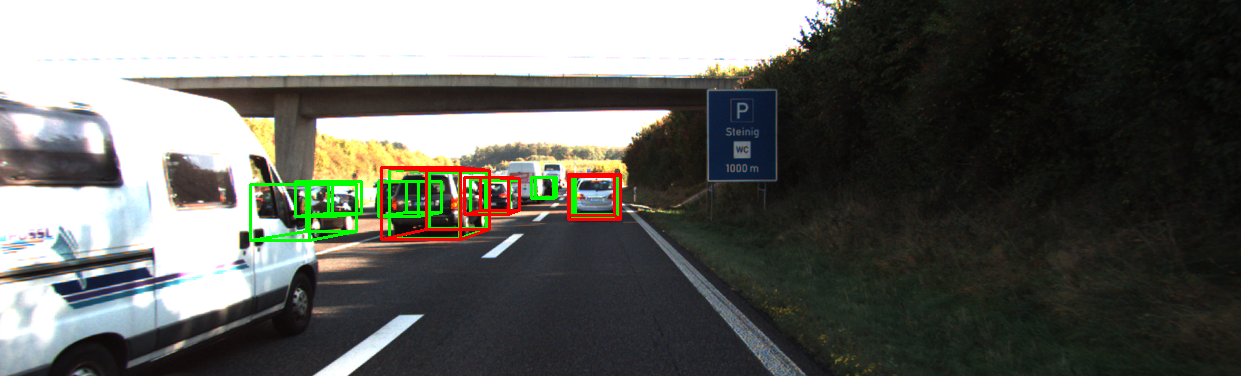} \hspace{-1.2em} & 
  \includegraphics[trim={0cm 0cm 0cm 0cm},clip,width=0.245\linewidth]{ 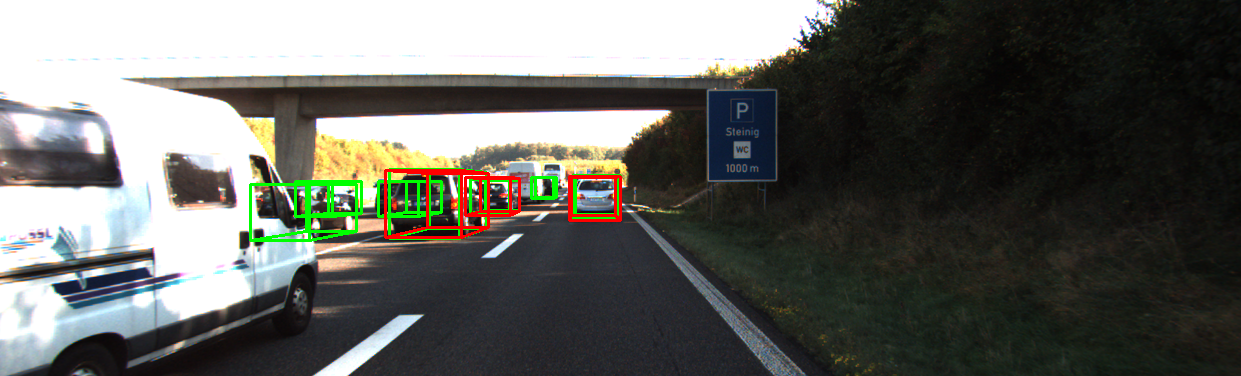} \hspace{-1.2em} & 
  
  \\

      \includegraphics[trim={0cm 0cm 0cm 0cm},clip,width=0.245\linewidth]{ 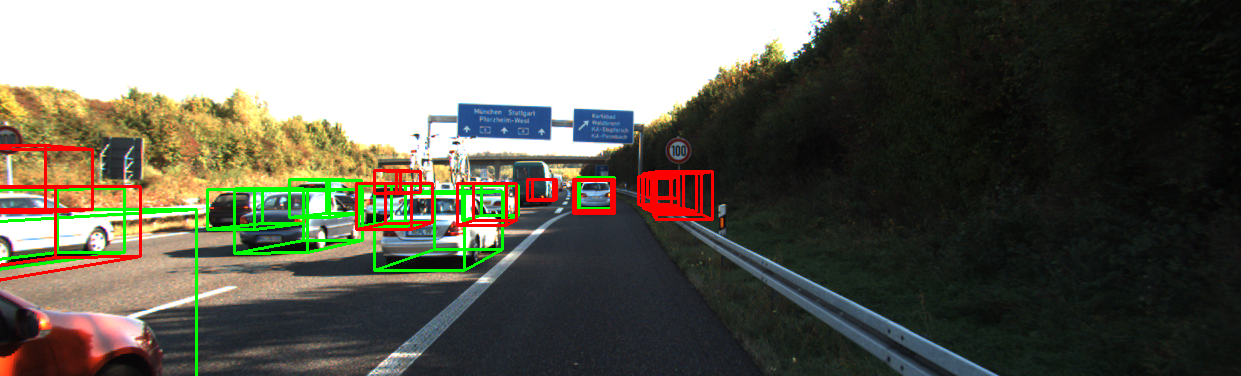} \hspace{-1.2em} &  
  \includegraphics[trim={0cm 0cm 0cm 0cm},clip,width=0.245\linewidth]{ figures/qual/waymo_second_sn_191.png} \hspace{-1.2em} & 
  \includegraphics[trim={0cm 0cm 0cm 0cm},clip,width=0.245\linewidth]{ figures/qual/waymo_second_sn_191.png} \hspace{-1.2em} & 
  \includegraphics[trim={0cm 0cm 0cm 0cm},clip,width=0.245\linewidth]{ 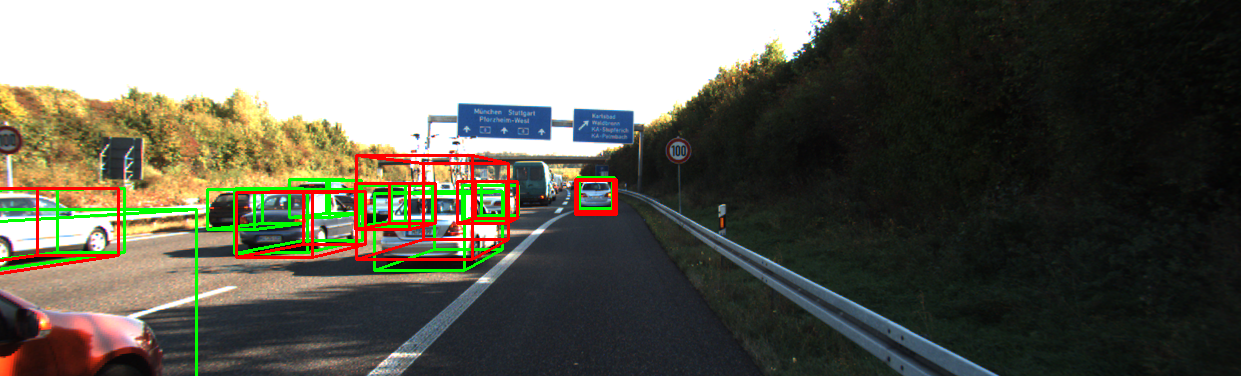} \hspace{-1.2em} & 
  
  \\
    
  Direct transfer \hspace{-1.2em} & ST \hspace{-1.2em}  & SN \cite{Wang2020TrainIG} \hspace{-1.2em} & Ours \hspace{-1.2em}
\end{tabular}
 	\vskip -10pt

\caption{A qualitative comparison of our bounding box predictions of the ``Car" class for the adaptation of SECOND-iou \cite{second} (rows 1-4) and PointRCNN \cite{shi2019pointrcnn} (rows 5-8), with direct transfer (DT), pseudo-label self training (ST), statistical normalization (SN) \cite{Cai2019ExploringOR}, and ST3D \cite{yang2021st3d}. Ground truth annotations are in green and predictions are in red. Best viewed zoomed in and in color.}
\label{fig:qual}
\vspace{-1em}
\label{port}
\end{figure*}

\begin{table*}[]
\centering
\caption{Ablation results. A comparison of 3D mAP performance of the adaptation of SECOND-iou using different prototype computation methods for two domain scenarios.}
\vspace{-1em}
\label{tab:proto}
\begin{tabular}{l|ccc|ccc}

\midrule[1pt]\toprule[0.1pt]
\multicolumn{1}{c|}{\multirow{2}{*}{Prototype computation method}} & \multicolumn{3}{c|}{Waymo $\rightarrow$ KITTI} & \multicolumn{3}{c}{nuScenes $\rightarrow$ KITTI} \\ \cline{2-7} 
\multicolumn{1}{c|}{}                                              & easy       & mod.       & hard      & easy     & mod.         & hard        \\ \hline
Average                                                           &  73.56          &     57.68        &   56.75        &     65.00    & 47.55    & 40.24   \\
Self attention                                                    & 72.90      & 56.82      & 55.21     & 65.01    & 47.54        & 42.43       \\
Transformer                                                       & 72.50      & 62.64      & 56.39     & 69.26    & 50.09        & 45.11       \\
Transformer + entropy weight                                      & 75.75     & 64.43      & 57.19     & 71.56    & 52.12        & 45.86       \\ \midrule[0.1pt]\toprule[1pt]
\end{tabular}
\vspace{-1.5em}
\end{table*}


\subsection{Comparison with state-of-the-art}
\noindent\textbf{Quantitative analysis.} We compare the mean average precision (mAP) of 3D bounding boxes across various difficulty settings for the domain shift scenarios Waymo $\rightarrow$ KITTI,  nuScenes$\rightarrow$ KITTI, KITTI $\rightarrow$ nuScenes, and Waymo$\rightarrow$ nuScenes. Where the target dataset is KITTI, we use the official evaluation metrics detailed in \cite{KITTI} with easy, moderate, and hard difficulty categories based on the distance and level of occlusion of the object from the sensor, with an IoU threshold of $0.7$. Where the target dataset is nuScenes, we use the metrics of \cite{nuscenes2019}, and average across difficulties to be consistent with both \cite{yang2021st3d} and \cite{Saltori2020SFUDA3DSU}. We reproduce the results of ST3D and SN using similar training lengths and batch sizes as our experiments. Note that due to computational limitations, the batch sizes are smaller than those used by \cite{yang2021st3d} and \cite{second}. Due to the lack of a code repository for SF-UDA\textsuperscript{3D} at the time of writing, we compare with the reported numbers.   We implement our method with SECOND-iou for comparison with ST3D and with PointRCNN for comparison with SF-UDA\textsuperscript{3D} to be consistent with their base object detector networks. This can be seen in Table \ref{tab:main-table}. We demonstrate the best results using both object detection networks in most categories, beating the weakly supervised approach of SN as well as \cite{Saltori2020SFUDA3DSU} and \cite{yang2021st3d}. We observe a $\boldsymbol{19.95\%}$ overall improvement from the closest performing competitor ST3D in the case of SECOND-iou in the nuScenes $\rightarrow$ KITTI domain scenario and a $\boldsymbol{5.76\%}$ 
improvement in the case of  Waymo $\rightarrow$ KITTI.

\noindent\textbf{Qualitative analysis.} We further demonstrate the effectiveness of our domain adaptation framework with those of recent methods through a visual comparison of the predicted bounding boxes for the Waymo $\rightarrow$ KITTI task for both object detection networks in Figure \ref{fig:qual}. The problems of direct transfer (DT) of the source model are localization and over-confident false positives (see column 1). This problem is mitigated only partially by pseudo labelling methods and by ST3D, and is better addressed by our proposed approach.
\begin{table}[]
\caption{Analyzing the effect of the iterative training scheme. A comparison of adaptation method 3D mAP performance at different meta-iterations for the Waymo $\rightarrow$ KITTI task.}
\vspace{-1em}
\label{tab:meta}
\begin{tabular}{c|c|ccc}
\midrule[1pt]\toprule[0.1pt]
\multirow{2}{*}{Domain shift}   & \multirow{2}{*}{\begin{tabular}[c]{@{}c@{}}meta \\ iteration\end{tabular}} & \multicolumn{3}{c}{mAP}                                                       \\ \cline{3-5} 
                                &                                                                            & \multicolumn{1}{l}{easy} & \multicolumn{1}{l}{mod.} & \multicolumn{1}{l}{hard} \\ \hline
\multirow{5}{*}{Waymo $\rightarrow$ KITTI} & DT                                                                         & 18.37                    & 17.31                    & 16.09                    \\
                                & 1                                                                          & 57.76                    & 43.02                    & 38.50                    \\
                                & 2                                                                          & 63.06                    & 46.71                    & 41.12                    \\
                                & 3                                                                          & 63.18                    & 46.93                    & 40.94                    \\
                                & 4                                                                          & 71.56                    & 52.12                    & 45.86                    \\ \midrule[0.1pt]\toprule[1pt]
\end{tabular}
\end{table}


\begin{figure}
    \vspace{-1em}
    \includegraphics[width=\linewidth]{ 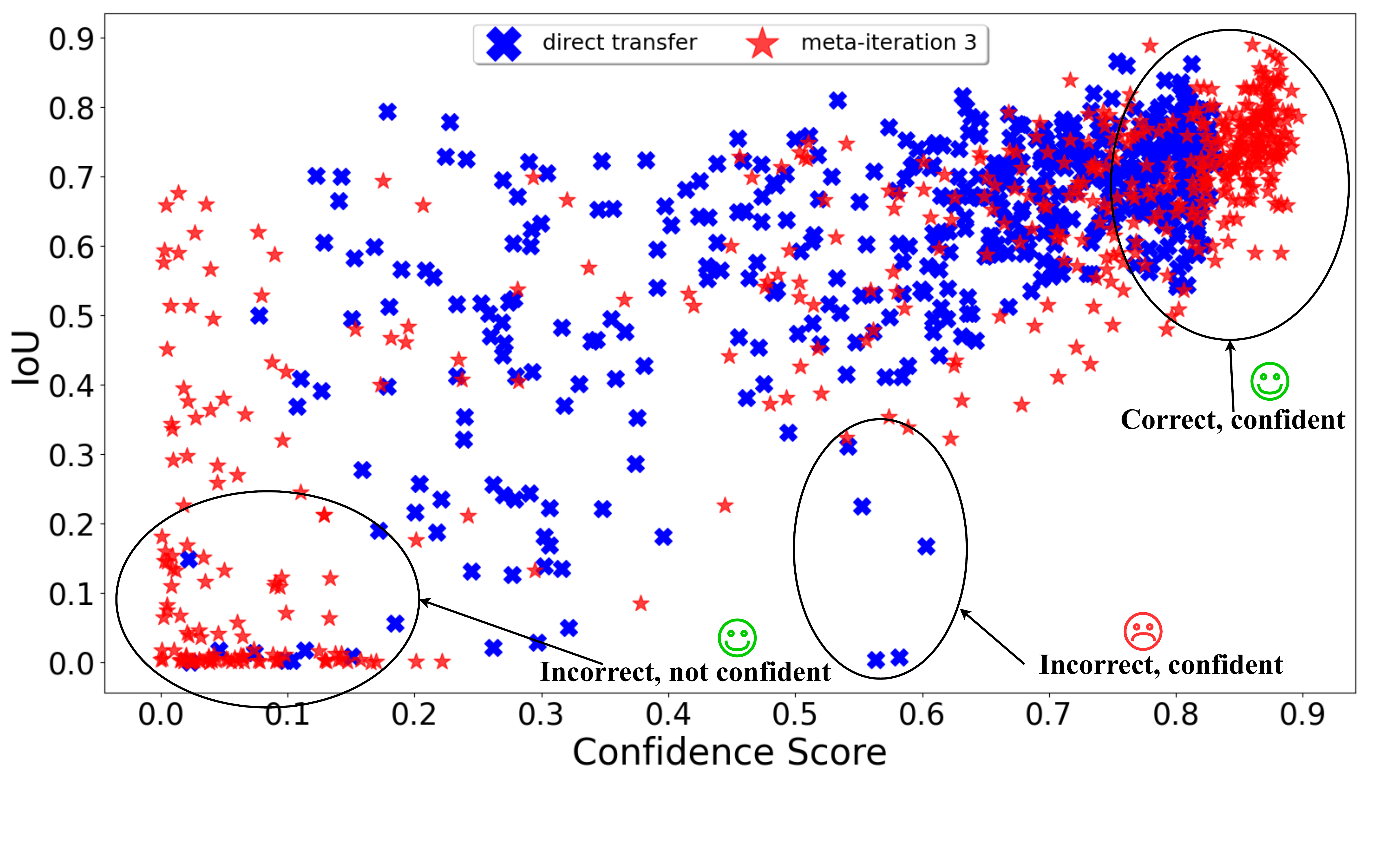}
    \vspace{-3em}
    \caption{A plot of IoU vs. confidence scores ($c$) of 500 pseudo labels of object detector \cite{second} at two stages of the iterative training process. At direct transfer(DT) from the source trained model (blue x's), several incorrect labels $IoU<0.5$ with the ground truth labels have confidence score $c >0.5$, while at meta-iteration 3 of the training scheme (red stars), most correct samples tend to have higher confidence than that of DT, and incorrect samples have $c <0.25$. This distribution being most dense at the lower left and upper right corners of the plot implies successive adaptation improves the quality of the pseudo labels. }
    \vspace{-1em}
    \label{fig:plot}
\end{figure}


\subsection{Ablation study}
We analyze the contribution of the different segments of our domain adaptation framework. In Table \ref{tab:proto}, we compare the 3D mAP performance of the network using four different prototype computation methods on two different domain shift scenarios for the SECOND-iou object detector; \textit{(i) Average}: Class prototype is computed by taking the mean of positive region features, \textit{(ii) Self-attention}: Prototype is computed by taking the mean of attentive region features, which are the result of sending region features to a single multi-head self attention block.  \textit{(iii) Transformer}: Prototype is computed by taking the mean of attentive region features, which are the result of sending region features to the transformer module detailed in Section \ref{sec:trans}. \textit{(iv.) Transformer with entropy weights}: Prototype is computed by taking the prediction entropy weighted mean of transformer generated attentive region features. This is the best performing approach.

Additionally, we compare the performance of the network at different meta-iterations of the training procedure in Table \ref{tab:meta}, observing improvements in performance with each successive meta-iteration. We further observe that this boost in precision saturates after a certain number of these meta-iterations. In order to visualize the quality of the pseudo labels at different stages of training, we plot the IoU with ground truth vs. confidence of 500 pseudo labels generated by the source only model (blue x's) and the third meta-iteration model (red stars). We desire that the model generate confident labels with large IoU scores (labels should be pushed to the upper right corner) and be uncertain about generated labels with low IoU scores (labels should be be pushed to the lower left). The labels generated after adaptation are less noisy than that of the source-generated labels, indicated by the fact that incorrect samples ($IoU<0.5$) of meta-iteration 3 tend to be distributed with lower confidence than that of DT labels. This shows that the quality of the pseudo-labels improves after adaptation.
\section{Limitations}
 We conduct experiments for a single class and  compute a single attentive prototype. The largest domain gaps often occur with smaller objects that appear less frequently such as pedestrians and cyclists. However, our method could be easily extended to learn multiple attentive prototypes for multi-object detection, with the use of class-specific transformer modules. The method is less effective when addressing shifts from smaller to larger datasets (Table \ref{tab:main-table}, KITTI to nuScenes). Additionally, in this work we perform closed-set domain adaptation, in which we make the assumption that the classes in the source domain have equivalent classes in the target domain. This may not hold true for all datasets in the general domain adaptation setting. 
\section{Conclusions and future work}
In this paper we proposed a source-free domain adaptation framework for unsupervised domain adaptive 3D object detectors that uses a transformer module to compute an attentive class prototype to perform pseudo-label refinement during self training. Our method outperforms other recent domain adaptation networks for several different domain shifts. We mainly address pseudo-label noise related to the false positive mis-classification of regions in the 3D scene, and not the dimensions of the bounding box. A factor that contributes the drop in performance upon domain shift is the difference in average size of the vehicles in different locations \cite{Wang2020TrainIG}. While they address this by the weakly supervised approach of statistical normalization, in the future we hope to provide a fully unsupervised solution.

{\small
\bibliographystyle{ieee_fullname}
\bibliography{egbib}
}

\end{document}